\documentclass[runningheads]{llncs}
\usepackage{graphicx}
\usepackage{tikz}
\usepackage{comment}
\usepackage{amsmath,amssymb} % define this before the line numbering.
\usepackage{color}

\usepackage[accsupp]{axessibility}  % Improves PDF readability for those with disabilities.

\usepackage{array}
\usepackage{multirow}
\usepackage{float}
\usepackage{stfloats}
\usepackage{wrapfig}
\usepackage{gensymb}
\usepackage{listings}
\usepackage{xcolor}

\usepackage[width=122mm,left=12mm,paperwidth=146mm,height=193mm,top=12mm,paperheight=217mm]{geometry}

\DeclareMathOperator*{\argmax}{argmax}
\def\ECCVSubNumber{0}  % Insert your submission number here

\begin{document}
% \linenumbers
\pagestyle{headings}
\mainmatter
\title{Weakly-Supervised End-to-End CAD Retrieval to Scan Objects} % Replace with your title

\titlerunning{Weakly-Supervised End-to-End CAD Retrieval to Scan Objects}
\author{Tim Beyer\and % \orcidID{0000-0003-0205-9578} \and
Angela Dai %\orcidID{0000-0002-6241-8782}
}
\authorrunning{T. Beyer et al.}
\institute{Technical University of Munich}
%\end{comment}
%******************

\maketitle
\begin{figure}[ht!]
    \vspace{-0.7cm}
    \centering
        \includegraphics[width=\linewidth]{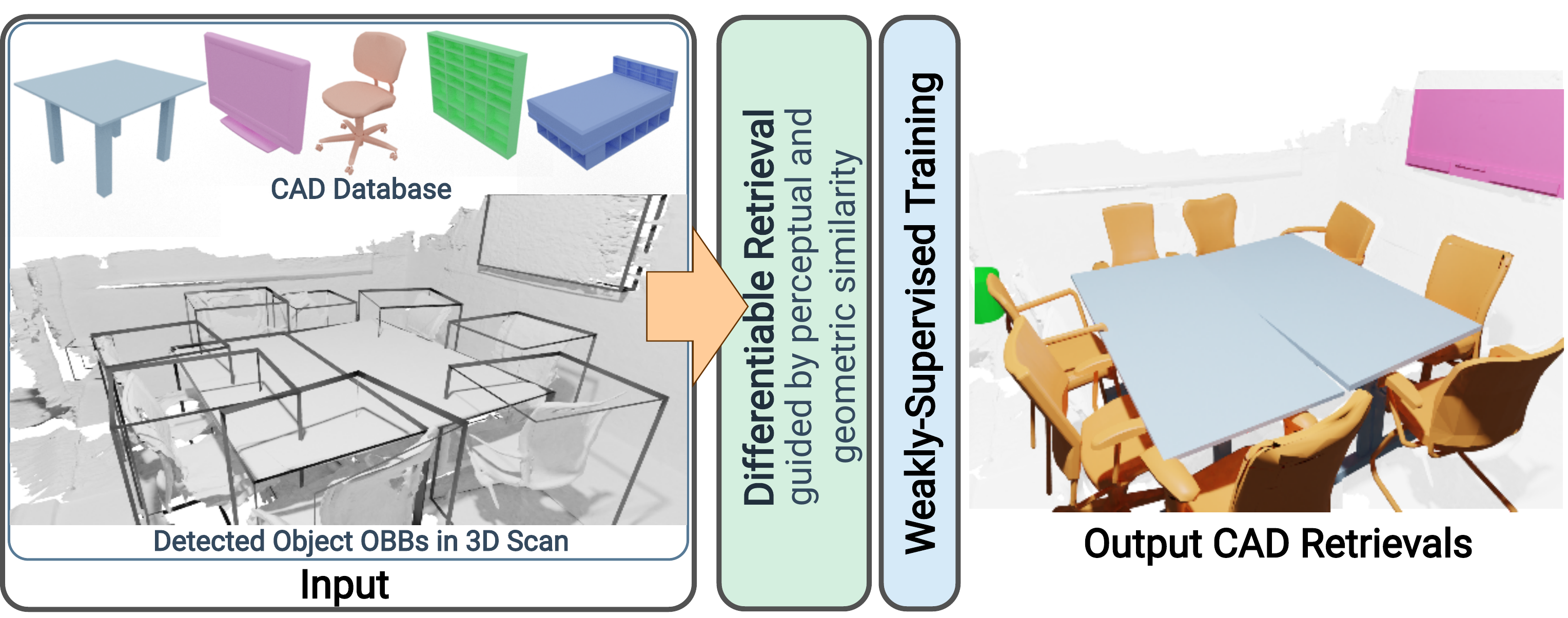}
    \label{fig:teaser}
    \vspace{-0.7cm}
       \caption{We present a new weakly-supervised approach for CAD model retrieval to objects detected in RGB-D scans.
       Given an input RGB-D scan of a scene and its object detections as oriented bounding boxes, along with a CAD database, we learn a joint embedding space of scan and CAD objects guided only by proxy geometric and perceptual similarities, without requiring any CAD-scan associations or annotations.
       Our weakly-supervised training through differentiable retrieval enables robust CAD retrieval, even for unseen class categories. 
       Using the alignment from the oriented bounding box detections, we can create a CAD-based reconstruction of the scanned scene.
       }
\end{figure}
\vspace{-1.0cm}

\vspace{-0.2cm}
\begin{abstract}
CAD model retrieval to real-world scene observations has shown strong promise as a basis for 3D perception of objects and a clean, lightweight mesh-based scene representation; however, current approaches to retrieve CAD models to a query scan rely on expensive manual annotations of 1:1 associations of CAD-scan objects, which typically contain strong lower-level geometric differences.
We thus propose a new weakly-supervised approach to retrieve semantically and structurally similar CAD models to a query 3D scanned scene without requiring any CAD-scan associations, and only object detection information as oriented bounding boxes.
Our approach leverages a fully-differentiable top-$k$ retrieval layer, enabling end-to-end training guided by geometric and perceptual similarity of the top retrieved CAD models to the scan queries.
We demonstrate that our weakly-supervised approach can outperform fully-supervised retrieval methods on challenging real-world ScanNet scans, and maintain robustness for unseen class categories, achieving significantly improved performance over fully-supervised state of the art in zero-shot CAD retrieval.
\end{abstract}

\section{Introduction}
The proliferation of commodity depth-sensing devices (e.g., Microsoft Kinect, Intel RealSense, recent iPhones) in recent years has driven remarkable advances in 3D capture and reconstruction of real-world environments. 
RGB-D reconstruction and tracking have enabled robust real-time capture in many real-world scenarios \cite{dai2017bundlefusion,izadi2011kinectfusion,niessner2013real,whelan2015elasticfusion}.
Unfortunately, the resulting reconstructed 3D models support only limited use in downstream applications, as they lack sharp details and remain incomplete, making them unsuitable for many tasks such as content creation or real-time robotics, in particular in comparison with clean, complete artist-created 3D models.

In contrast, digitizing real-world reconstructions to 3D CAD representations of the observed objects enables a lightweight 3D mesh representation that can directly be consumed by downstream application pipelines as well as creating insights towards bridging a synthetic-real domain gap under limited 3D data scenarios.
This remains a challenging problem due to the strong lower-level geometric differences between synthetic CAD models and real-scanned objects, posing notable difficulties for handcrafted feature correspondence.
With recent advances in data-driven machine learning, the simultaneous availability of large-scale synthetic 3D CAD data \cite{chang2015shapenet,fu20213d} along with real-world RGB-D scans of scenes \cite{matterport3d,dai2017scannet} has paved the way towards learning to transform real-world observations to clean, compact CAD-based representations of the observed objects in a scene.
Thus, several recent works have established manual CAD retrieval and alignment annotations to real-world RGB-D scans \cite{avetisyan2019scan2cad,dahnert2019joint} to develop data-driven learning methods for CAD retrieval and alignment to real-world RGB-D scans and even single RGB images \cite{kuo2020mask2cad,kuo2021patch2cad,maninis2022vid2cad}.
While these methods show a strong promise in CAD-based 3D perception, they all rely on supervision from an expensive manual annotation process to retrieve and align CAD models to real observations where exact synthetic-real matches cannot be expected to exist in a database.

Thus, we propose to instead learn to retrieve similar CAD models to real-world scanned objects in an weakly-supervised, end-to-end fashion. 
From an input RGB-D scan with oriented bounding box object detections, we retrieve for each detected object the most similar CAD model from a database, allowing for complete and compact object-based 3D reconstruction of scenes from noisy and incomplete scans. 
To this end, we propose a differentiable top-$k$ search for similar CAD models to real object scans.
Based on predicted geometric and perceptual similarity between CAD and scan objects, we learn a structured space where similar CAD and scan objects lie together regardless of the source domain, trained end-to-end without requiring any manual CAD-scan annotations.
At test time, we can thus retrieve similar CAD models for detected 3D scan objects via a nearest neighbor search in our learned embedding space.
Furthermore, our weakly-supervised training paradigm enables robust CAD retrieval even on new class categories.
\newpage
In summary, our contributions are:
\begin{itemize}
    \item We present a new, weakly-supervised approach for CAD model retrieval to real-world scans that avoids requiring expensive CAD model annotation to real observations.
    \item We propose a differentiable top-$k$ approach to enable robust CAD retrieval to geometrically and perceptually match query 3D scans of objects
    \item Our weakly-supervised approach, given only predicted object box detections in a scanned scene, achieves a relative improvement of 12\% in top-1 CAD retrieval accuracy over the fully supervised state of the art on challenging real-world ScanNet data and demonstrates large improvements on unseen classes, retrieving the correct object 175\% more often than existing methods.
\end{itemize}

\section{Related work}
\paragraph{3D Shape Descriptors}
Shape analysis with descriptors has seen a long history in geometry understanding, with many applications towards correspondence-finding, matching, and retrieval.
Many shape descriptors have been developed as hand-crafted features based on a suite of local geometric characteristics such as normals, curvature, distances \cite{gelfand2005robust,ohbuchi2005shape,osada2002shape,rusu2009fast,tombari2010unique} as well as some higher-level features such as topology \cite{hilaga2001topology,sundar2003skeleton}.
Recent advances in deep learning on 3D volumes \cite{qi2016volumetric,zeng20173dmatch} and points \cite{qi2017pointnet,qi2017pointnet++} have enabled training neural networks for correspondence-finding and classification tasks to provide feature descriptors for 3D shapes.
For such shape descriptors, the CAD retrieval task remains a significant challenge due to the strong low-level geometric differences between noisy, incomplete real scans and clean, uncluttered CAD models.

\paragraph{3D Reconstruction with CAD Model Retrieval} 
With the introduction of large-scale synthetic 3D shape datasets such as ShapeNet~\cite{chang2015shapenet}, various approaches have been proposed to retrieve and align CAD models from a database to produce an object-based reconstruction from real-world RGB-D observations. 
Earlier approaches employed hand-crafted features to ground learning or correspondence finding between template or CAD models and scanned observations~\cite{kim2012acquiring,li2015database,nan2012search}.
Following the introduction of large-scale real-world scan datasets~\cite{matterport3d,dai2017scannet}, Avetisyan et al. introduced an annotated dataset of CAD retrieval and alignment to RGB-D scans~\cite{avetisyan2019scan2cad}, along with a deep network to establish correspondences across the domains to solve for pose alignment of CAD models~\cite{avetisyan2019end}.
Several recent approaches have also tackled CAD model retrieval and alignment to image observations by retrieving rendered views of CAD models based on learned feature similarities~\cite{huang2018holistic,izadinia2017im2cad,kuo2020mask2cad,kuo2021patch2cad}.
Deformation-based approaches following CAD retrieval have also been proposed to allow database models to better fit to observed input geometry~\cite{ishimtsev2020cad,uy2020deformation}.
Our approach is inspired by the success of CAD model retrieval and alignment to real-world observations, but we focus on effective CAD retrieval to scans without requiring any CAD-scan correspondence association, as has been used in these previous methods.

\paragraph{Learned 3D Representations with Multi-modal Data}
Several works have proposed to learn representations across different data modalities to establish ties between 3D geometry and images \cite{hou2021pri3d,li2015joint} and also language \cite{herzog2015lesss,Wu2021Towers}.
Recent works have also employed constrastive learning~\cite{van2018representation} to construct joint embedding spaces between 3D CAD models and real-world RGB or RGB-D image observations to enable CAD retrieval to image queries \cite{kuo2020mask2cad,kuo2021patch2cad,zou2019complete}, using supervised CAD-scan correspondence information to inform positive correspondences.
Similar to our setting, Dahnert et al.~\cite{dahnert2019joint} recently proposed to learn a joint embedding space between 3D CAD models and scan objects, constructed with a triplet loss formulation based on supervised CAD-scan association annotations.
In contrast to these approaches that leverage correspondence annotations of CAD models to real-world observations, we propose to learn a weakly-supervised fully differentiable retrieval without requiring any expensive CAD-scan association annotations.

\paragraph{Differentiable Top-K}
Many retrieval systems characterize retrieval as a nearest-neighbor lookup into an embedding space from a query embedding vector.
The standard nearest neighbor selection process is not differentiable; thus many approaches instead construct differentiable surrogate learning objectives such as a triplet loss that do not directly represent inference behavior performance.
Recently, several methods to construct differentiable top-$k$ operations were proposed, most notably optimal transport-based methods \cite{cuturi2019differentiable}, Sinkhorn operators \cite{xie2020differentiable} and an extension of the Gumbel-Softmax trick \cite{xie2019reparameterizable}. 
In this work, we employ a formulation based on perturbed optimizers \cite{berthet2020learning} that was first proposed for image patch selection for a classification task \cite{cordonnier2021differentiable}.

\section{Method}

\subsection{Overview}

\begin{figure}
\centering
\includegraphics[width=\textwidth]{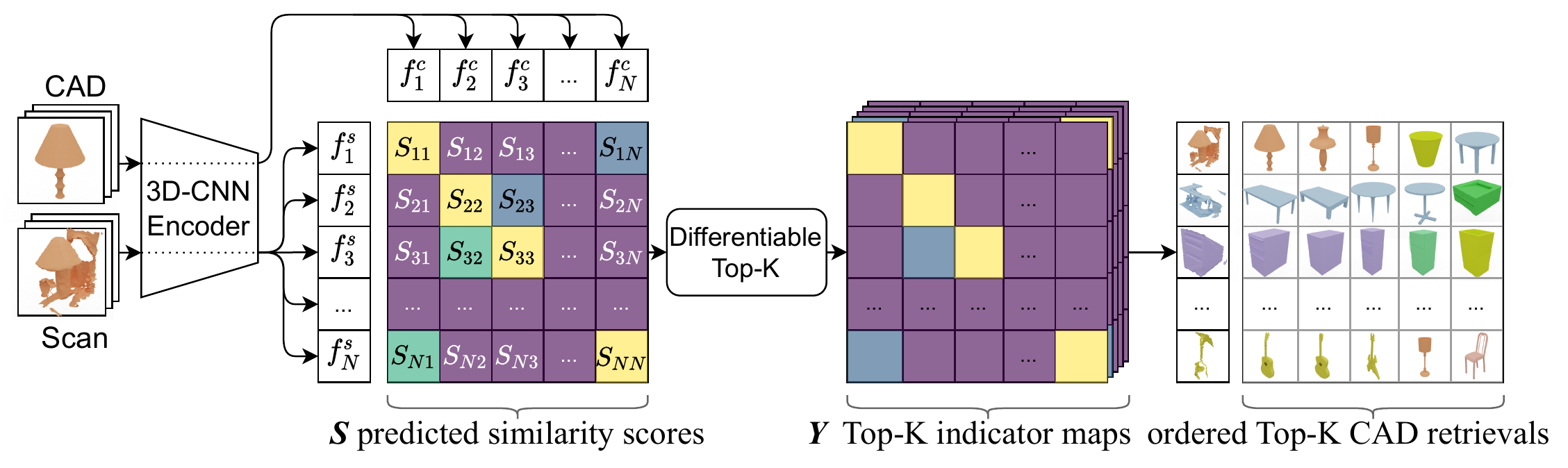}
\caption{Detected scan objects and CAD models are encoded with a siamese 3D CNN to map to 128-dimensional normalized embeddings $\{f_i^s\}$ and $\{f_i^c\}$. 
The similarity of each pair of scan and CAD objects \textit{\textbf{S}} is then computed as the cosine similarities of their embedding features. 
These similarities are employed in our differentiable top-$k$ retrieval layer across the scan rows to obtain for each scan $k$ soft indicator vectors over the CAD models that sum to 1.
In the absence of ground truth scan-CAD associations, we guide the indicator estimation by perceptual and geometric similarity of each potential CAD model in reconstructing the scan query.
}
\label{fig:overview}
\end{figure}
Our method learns to map together detected real-world scanned objects and synthetic CAD models into a shared embedding space, such that for a given scan object, similarly structured CAD models lie near the scan, regardless of lower-level geometric differences like noise, incompleteness, and background clutter.
This then enables finding the most similar CAD model for a query detected scan object by nearest neighbor retrieval.
Crucially, we aim to learn this embedding space without requiring any knowledge of scan-CAD associations, as such data annotations require an expensive manual process.
Instead, we consider during training only a CAD model database along with a set of RGB-D scanned scenes that have object box annotations available for object detection supervision (much cheaper annotations than CAD model retrieval to scans, which requires annotator search through the large database to find the most perceptually similar model).
At test time, for each detected object in a new RGB-D scan, we can use its embedding feature to retrieve the most similar CAD model from the database to represent that object.

An overview of our weakly-supervised embedding learning is shown in Figure~\ref{fig:overview}.
Detected scan objects and CAD models are both represented as volumetric occupancy grids and encoded into the joint embedding space with a siamese 3D convolutional neural network. %(Sec.~\ref{sec:encoder-architecture}).
We then propose a differentiable retrieval, guided by the geometric and perceptual similarity of the scan and CAD objects. % (Sec.~\ref{sec:diff-top-k}).
This enables robust CAD retrieval to new scan objects at test time, without requiring any knowledge of 1:1 scan-CAD correspondences.

\subsection{Scan and CAD Encoding}
\label{sec:encoder-architecture}
Scan and CAD objects $\{s_i\}$ and $\{c_i\}$, respectively, are encoded by a 3D convolutional neural network with residual connections that takes as input a $32^3$ occupancy grid and produces a 128-dimensional embedding, resulting in normalized encodings $\{f_i^s\}$ and $\{f_i^c\}$.
The encodings of $\{s_i\}$ and $\{c_i\}$ are constructed in siamese fashion with the same 3D CNN weights.
Since the $32^3$ occupancy grid object representation lacks the objects' scale, we additionally pass the 3-dimensional scale to our model by concatenating it to the feature vector before the final fully connected layer.

From the encodings $\{f_i^s\}$ and $\{f_i^c\}$, we construct a pairwise cosine similarity matrix $\mathbf{S}$ describing scan-CAD similarities as $\mathbf{S_{ij}}=f_i^s\cdot f_j^c$.
A differentiable top-$k$ retrieval layer then guides the construction of the embedding space by enabling gradient propagation through the selection of the top-$k$ most similar CAD models.

\newpage
\subsection{Differentiable Top-$k$ Retrieval}
\label{sec:diff-top-k}
To learn an effective CAD-scan embedding space, we formulate a differentiable top-$k$ retrieval for CAD models to scan object queries.
The top-$k$ operator accepts the pairwise cosine similarity scores $\mathbf{S}$ of scan and CAD encodings and produces indicator vectors $\mathbf{Y}$ to the top-$k$ items in each row of $\mathbf{S}$ (see Figure~\ref{fig:overview}). 
At test time, $\mathbf{Y}$ can be computed with a hard top-$k$ operator to select the $k$ most similar CAD models to each scan. 

Our top-$k$ formulation is inspired by Cordonnier et al.~\cite{cordonnier2021differentiable} who proposed a differentiable image patch selection for classification tasks based on the perturbed maximum method \cite{berthet2020learning}.
The perturbed maximum method can make any linear program of the form \begin{equation}
    \argmax_{\mathbf{Y}\in C} \langle\mathbf{Y},\mathbf{S} \rangle
    \label{eq:linear-program}
\end{equation}
differentiable, with convex polytope constraint set $C$, variables to be optimized $\mathbf{Y}$ and inputs $\mathbf{S}$. 
Given a linear program of this form, the differentiable forward pass is computed by taking the expectation over multiple perturbed inputs $\mathbf{S}$:
\begin{equation}
    \mathbf{Y}_\sigma = \mathbb{E}_\mathbb{Z} \left[{\argmax_{\mathbf{Y}\in C} \langle\mathbf{Y},\mathbf{S} + \sigma \mathbf{Z} \rangle.}\right] \; \text{where} \;\; \mathbf{Z} \sim \mathcal{N}(0, 1),
\end{equation}
where $\sigma$ and the number of samples $n$ over which to compute the expectation are hyper-parameters.
We use $\sigma=0.05$ and $n=1000$ for all experiments unless specified otherwise.
For Gaussian noise $\mathbf{Z}$ the backward pass is given as:
\begin{equation}
    J_{\mathbf{S}}\mathbf{Y}_\sigma = \mathbb{E}_\mathbb{Z} \left[{\argmax_{\mathbf{Y}\in C} \langle\mathbf{Y},\mathbf{S} + \sigma \mathbf{Z} \rangle \mathbf{Z}^\intercal/\sigma}\right].
\end{equation}
The top-$k$ operation can be cast as a linear program that conforms to Eq.~\ref{eq:linear-program}:
\begin{equation}
    \max_{\mathbf{Y}\in C} \langle \mathbf{Y},\mathbf{S} \mathbf{1}^\intercal \rangle.
\end{equation}
with the constraint set
\begin{equation}
C = \mathbf{Y} \in \mathbb{R}^{N\times K}: \mathbf{Y}_{n,k} \le 0, \mathbf{1}^\intercal \mathbf{Y}=\mathbf{1},\mathbf{Y}\mathbf{1}\le \mathbf{1}, \sum_{i\in[N]} i\mathbf{Y}_{i,k} < \sum_{j\in[N]} j\mathbf{Y}_{j,k'} \forall k < k'
\end{equation}
The four constraints are responsible for ensuring that $\mathbf{Y}$ is a valid distribution with each column having a total weight of 1. 
The final condition ensures that all indices are sorted.
Since the inputs to our top-$k$ are the intra-batch embedding cosine similarities, they are already normalized into the range $[0, 1]$, however in principle any score values could be used as the input without normalization.

\paragraph{Loss formulation.}
\label{sec:losses}
We thus consider the task of retrieving the top $k$ CAD models $\mathbf{R}$ with the highest similarity to the query scan $Q$, in an end-to-end fashion:
\begin{equation}
    \argmax_{\textbf{R}} \; \text{Score}(Q, \mathbf{R}) = \frac{1}{K} \sum_{k=1}^{K} \text{Similarity}(Q, \mathbf{R}_{k}) \label{eq:retrieval-score}
\end{equation}
By employing the differentiable top-$k$ layer, we can differentiate through this function during training.
Given the predicted pairwise similarity scores $\mathbf{S}$, we obtain a set of $k$ (nearly) one-hot indicator vectors $\mathbf{Y_i} = \text{DiffTopK}(\mathbf{S_{i}})$, which are the indices of the retrievals \textbf{R} for the $i^{\textrm{th}}$ scan.
Without ground truth CAD correspondences for scanned objects, we instead estimate the geometric and perceptual similarity of potential CAD models as reconstructions of each scan query, computed by $\mathbf{P}_{ij}$ between scan query $i$ and CAD object $j$; see Sec.~\ref{sec:losses} for a full description.
We then aim to minimize the loss:
\begin{equation}
    \text{Loss}(\mathbf{S_i}, \mathbf{P_i}) = - \frac{1}{K} \sum_{j=1}^{N_{cads}}  \sum_{k=1}^{K} \mathbf{Y_{ijk}} \cdot \mathbf{P_{ij}}
    \label{eq:loss-single-instance-unordered}
\end{equation}
Note that Eq.~\ref{eq:retrieval-score} is maximized when Eq.~\ref{eq:loss-single-instance-unordered} is minimized. 

\paragraph{Ordered top-$k$ retrieval.}
This formulation, however, does not impose an ordering on the top-$k$ retrievals, whereas in the CAD retrieval scenario we would like to return an ordered top-$k$ set of nearest neighbor retrievals for a query. 
We thus multiply \textbf{P} with soft top-$k$ indicators, such that the top-$k$-th retrieval focuses on the $k^{\textrm{th}}$ most similar target.
The soft top-$k$ operator is derived from the differentiable top-$k$ layer, however we only require the forward pass and set $\sigma=0.005$.
$$\mathbf{\hat{P}} = \text{SoftTopK}(\mathbf{P}) \cdot \mathbf{P}$$
Our final loss function for one batch compares the predicted similarity matrix \textbf{S} with the estimated proxy retrieved CAD-scan similarity \textbf{P} as follows: 
\begin{equation}
\begin{split}
    \text{Loss}(\mathbf{S}, \mathbf{P}) &= - \frac{1}{N_{scans}} \cdot \frac{1}{K}  \sum_{i=1}^{N_{scans}} \sum_{j=1}^{N_{cads}} \sum_{k=1}^{K}  \mathbf{Y_{ijk}} \cdot \mathbf{\hat{P}_{ijk}}, \\
    \text{where} \;\;\; \mathbf{Y_i} & = \text{DiffTopK}(\mathbf{S_{i}}) 
    \;\;\; \text{and} \;\;\; \mathbf{\hat{P}} = \text{SoftTopK}(\mathbf{P}) \cdot \mathbf{P}.
\end{split}
\end{equation}

\begin{comment}
Each training batch consists of 64 randomly sampled scans and their top 1 most similar cad model as judged by the proxy metric, the loss is computed across the entire batch.
Traditional contrastive losses operate on the embedding space which is not a direct formulation. 
In retrieval systems, the embeddings themselves are not of immediate interest, only the retrieved instances. 
Therefore, we cast the object retrieval task as ``Retrieve the Top-K objects $\mathbf{R}$ with the highest similarity to the query scan \textit{Q}", a formulation that is 
1) directly aligned with our definition of retrieval quality, % NOTE: rephrase 
2) end-to-end differentiable and therefore, 
3) trainable.
For a single scan instance, we can formalize the problem as 
\begin{equation}
    \argmax_{\textbf{R}} \; \text{Score}(Q, \mathbf{R}) = \frac{1}{K} \sum_{k=1}^{K} \text{Similarity}(Q, \mathbf{R}_{k}) \label{eq:retrieval-score}
\end{equation}
\end{comment}

\subsection{Geometric- and Perceptually-Guided CAD Retrieval}
\label{sec:geom}
In the absence of any scan-CAD association information, we instead look to guide the similarity prediction by the geometric and perceptual quality of a retrieved CAD as a reconstruction of a scan query.
Since scan and CAD objects share higher-level similarities in semantics and structure, but differ noticeably in lower-level geometric characteristics, we characterize perceptual reconstruction similarity by rendering object views of scan and CAD and evaluate their similarity $F_{percep}$ based on the cosine similarity of their VGG-19~\cite{simonyan2014very} feature maps.
To perceptually compare a potential retrieved CAD model to a scan query, we render the depth and normals as $128\times128$ images that are composited together to form $I^s_k$ and $I^c_k$ for scan and CAD, respectively.
For a detailed description of the compositing procedure, we refer to the appendix.
We then estimate the perceptual similarity of $I^s_k$ and $I^c_k$ as the cosine similarity of their VGG-19~\cite{simonyan2014very} feature maps after the penultimate max-pooling layer.

For each potential retrieved CAD for a scan query, we render $N_{\textrm{view}}$ different views, and average the cosine similarities over all views, weighted by the proportion of pixels that are occupied by the scan object (as the scan is often incomplete whereas CAD models are always complete).

To additionally capture global geometric similarity, we consider the masked intersection-over-union $F_{geo}$ between a potential retrieved CAD for its scan query, which computes IoU only for regions in the scan that have been observed by a camera (so as to not penalize for scan incompleteness). 
Before computing $F_{geo}$, we anisotropically re-scale the CAD models to fit the aspect ratio of the detected box of the scan query to achieve better overlap.

We thus finally use $\mathbf{P}_{ij} = w_{percep}F_{percep}(\mathbf{S}_i, \mathbf{C}_j) + w_{geo}F_{geo}(\mathbf{S}_i, \mathbf{C}_j)$.

\subsection{Implementation details}
\label{sec:training-detail}
We train our system end-to-end using the Adam optimizer with batch size 64. 
We use a constant learning rate of $0.0003$ and train for 50 epochs until convergence.
Our perceptual and geometric similarity estimates (c.f. Section~\ref{sec:geom}) are combined to measure CAD-scan similarity using weights $w_{percep}=0.7$ and $w_{geo}=0.3$.
For each batch, we randomly select 64 scan objects and their top-1 matching CAD object as judged by the proxy metric; no further sampling strategy is used.
To estimate perceptual similarity, we render 5 views of each object using a fast renderer \cite{ravi2020pytorch3d}, such that the entire object is visible in the image.
For additional details on view selection for rendering, we refer to the appendix.

All scan and CAD object geometry are represented as $32^3$ binary voxel occupancy grids. 
We additionally add 2 voxels of padding to each side to account for potential small errors in object detection.
For evaluation on predicted oriented bounding boxes we employ CanonicalVoting~\cite{you2021canonical}.
The visibility information used for the geometric masked IoU measure is determined by volumetric fusion~\cite{curless1996volumetric} of depth frames in the scan to create a signed distance function; positive values indicate observed space, while negative values represent unobserved space.

\section{Results}
We evaluate our weakly-supervised, differentiable shape retrieval for CAD model retrieval to scan object queries from detected objects in ScanNet~\cite{dai2017scannet} scans.
Our approach enables a robust understanding of scan-CAD object similarities, enabling effective CAD retrieval on both seen and unseen class categories.

\subsubsection{Data}
For scan queries, we consider scanned objects from ScanNet~\cite{dai2017scannet} scans, with ShapeNet~\cite{chang2015shapenet} models as our CAD database.
All database CAD models are centered and normalized such that the bounding box diagonal has length 1.
We evaluate retrieval to scan objects detected with both ground truth bounding boxes as well as from predicted oriented bounding boxes from CanonicalVoting~\cite{you2021canonical}.
We then employ two datasets for evaluation on seen and unseen classes, respectively: 
\begin{itemize}
    \item \textit{Scan-CAD Object Similarity Dataset (seen classes):} has been established by Dahnert et al.~\cite{dahnert2019joint} as a benchmark for fine-grained CAD retrieval, with detailed scan-CAD rankings for 5,102 scan queries. Due to this limited number of annotations, with various uncommon categories having only single-digit number of annotations and thus of unreliable statistical significance, we adopt the original dataset splits and inference setting proposed by Dahnert et al.~\cite{dahnert2019joint}, thus evaluating on classes that have all been seen during training.
    \item \textit{Scan2CAD (unseen classes):} \cite{avetisyan2019scan2cad} contains 14,225 annotated CAD models to scanned scenes but without any detailed ranking annotations. The much larger set of annotations enables robust evaluation on  rare and unseen classes.
    We train on 8 categories and evaluate on 5 unseen categories, which produces 9564/2750/1077 train/val/test scan queries.
    Our CAD database at test time contains the 3049 ShapeNet models from Scan2CAD.
\end{itemize}
For additional dataset details and class category breakdowns, we refer to the appendix.

\subsubsection{Evaluation Metrics}
To evaluate our CAD retrieval method, we evaluate both retrieval and ranking quality as well as geometric similarity of the retrieved CAD to the ground truth CAD annotation.
We thus consider:
\begin{itemize}
  \item \textit{Top-1 retrieval accuracy (Top-1)}: measures whether the top-1 retrieved model is in the set of annotated matching CAD models to the scan (which can be up to 3 CADs in the Scan-CAD Object Similarity Dataset).
  \item \textit{Top-5 retrieval accuracy (Top-5)}: measures whether the ground truth CAD model is in the top 5 retrieved models.
  \item \textit{IoU (Top-1)}: evaluates the IoU of the $32^3$ geometric occupancy of the top-1 retrieved CAD model with the ground truth CAD annotation, which gives a measure of geometric similarity independent of the binary accuracy of retrieval correctness.
  \item \textit{IoU (Top-5)}: evaluates the average IoU of the top-5 retrieved CAD models with the ground truth CAD annotation.
  \item \textit{Category accuracy (Cat)}: is a coarse evaluation measure that follows prior retrieval works \cite{hua2017shrec,qi2016volumetric} to consider whether the top-1 retrieval has the same class as the scan object query.
  \item \textit{Ranking quality (RQ)}: measures the proportion of model retrievals that are retrieved with the correct rank; we evaluate this only when ranking annotations are available. 
  \item \textit{Mean reciprocal rank (MRR)}: is a metric commonly used in retrieval and recommender systems, and is defined as the average reciprocal rank of the ground truth CAD annotation \cite{radev2002evaluating,voorhees2001overview}. 
\end{itemize}
For each metric, we report the average over all instances; per-class performance evaluation is included in the appendix.

\newcommand\imgsizeje{.087\textwidth}

\begin{figure}[t]
  \centering
  \begin{tabular}{| c | c | c | c | c | c | c | c | c | c | c |}
    \hline
    \rotatebox[origin=c]{90}{Query}
    &
    \begin{minipage}{\imgsizeje}
      \includegraphics[width=\linewidth]{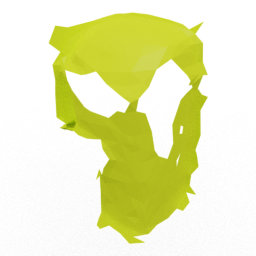}
    \end{minipage}
    &
    \begin{minipage}{\imgsizeje}
      \includegraphics[width=\linewidth]{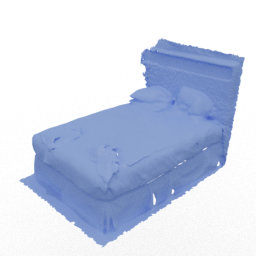}
    \end{minipage}
    &
    \begin{minipage}{\imgsizeje}
      \includegraphics[width=\linewidth]{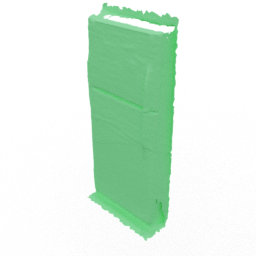}
    \end{minipage}
    & 
    \begin{minipage}{\imgsizeje}
      \includegraphics[width=\linewidth]{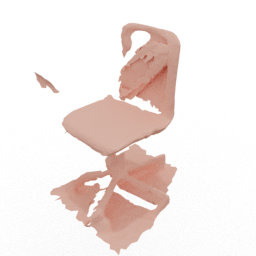}
    \end{minipage}
    & 
    \begin{minipage}{\imgsizeje}
      \includegraphics[width=\linewidth]{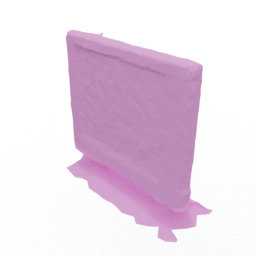}
    \end{minipage}
    & 
    \begin{minipage}{\imgsizeje}
      \includegraphics[width=\linewidth]{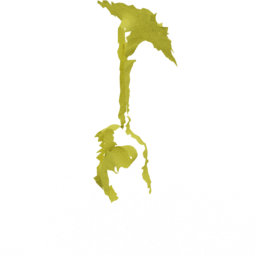}
    \end{minipage}
    & 
    \begin{minipage}{\imgsizeje}
      \includegraphics[width=\linewidth]{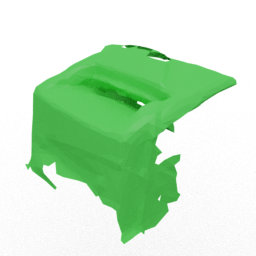}
    \end{minipage}
    & 
    \begin{minipage}{\imgsizeje}
      \includegraphics[width=\linewidth]{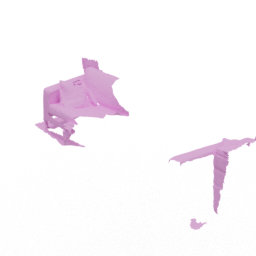}
    \end{minipage}
    & 
    \begin{minipage}{\imgsizeje}
      \includegraphics[width=\linewidth]{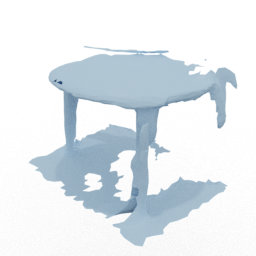}
    \end{minipage}
    & 
    \begin{minipage}{\imgsizeje}
      \includegraphics[width=\linewidth]{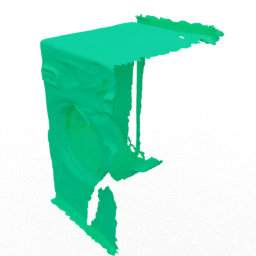}
    \end{minipage}
    \\ \hline \hline
    \rotatebox[origin=c]{90}{FPFH}&
    \begin{minipage}{\imgsizeje}
      \includegraphics[width=\linewidth]{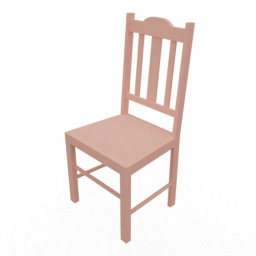}
    \end{minipage}
    &
    \begin{minipage}{\imgsizeje}
      \includegraphics[width=\linewidth]{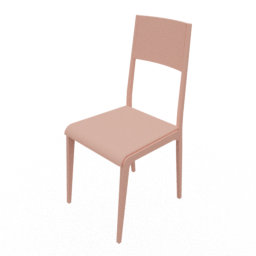}
    \end{minipage}
    & 
    \begin{minipage}{\imgsizeje}
      \includegraphics[width=\linewidth]{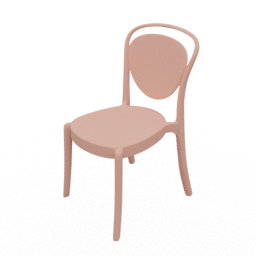}
    \end{minipage}
    & 
    \begin{minipage}{\imgsizeje}
      \includegraphics[width=\linewidth]{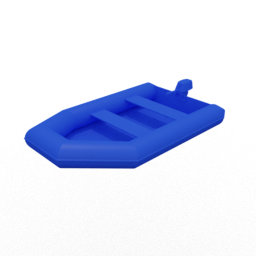}
    \end{minipage} 
    & 
    \begin{minipage}{\imgsizeje}
      \includegraphics[width=\linewidth]{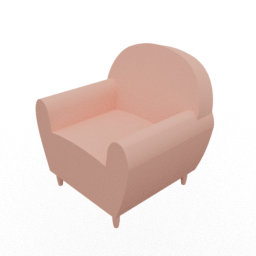}
    \end{minipage}
    & 
    \begin{minipage}{\imgsizeje}
      \includegraphics[width=\linewidth]{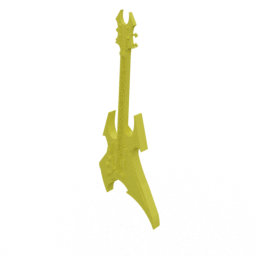}
    \end{minipage}
    & 
    \begin{minipage}{\imgsizeje}
      \includegraphics[width=\linewidth]{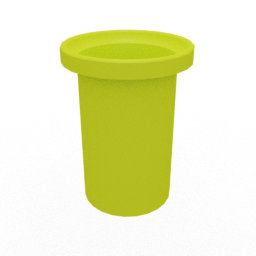}
    \end{minipage}
    & 
    \begin{minipage}{\imgsizeje}
      \includegraphics[width=\linewidth]{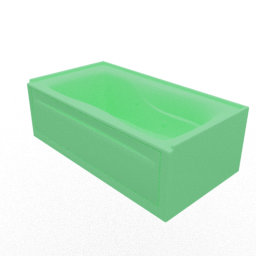}
    \end{minipage}
    & 
    \begin{minipage}{\imgsizeje}
      \includegraphics[width=\linewidth]{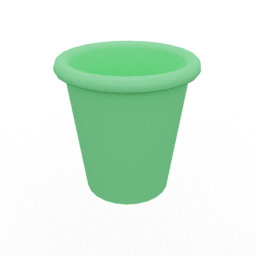}
    \end{minipage}
    & 
    \begin{minipage}{\imgsizeje}
      \includegraphics[width=\linewidth]{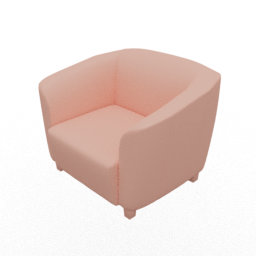}
    \end{minipage}
    \\ \hline  
    \rotatebox[origin=c]{90}{PointNet}&
    \begin{minipage}{\imgsizeje}
      \includegraphics[width=\linewidth]{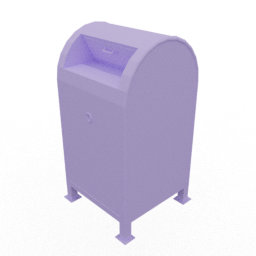}
    \end{minipage}
    &
    \begin{minipage}{\imgsizeje}
      \includegraphics[width=\linewidth]{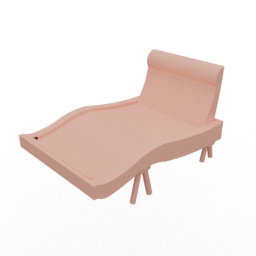}
    \end{minipage}
    & 
    \begin{minipage}{\imgsizeje}
      \includegraphics[width=\linewidth]{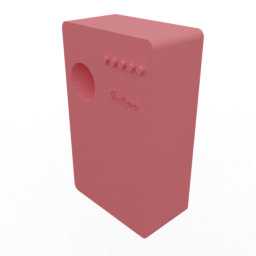}
    \end{minipage}
    & 
    \begin{minipage}{\imgsizeje}
      \includegraphics[width=\linewidth]{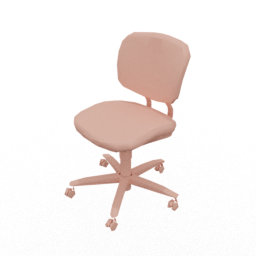}
    \end{minipage}
    & 
    \begin{minipage}{\imgsizeje}
      \includegraphics[width=\linewidth]{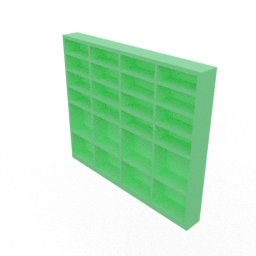}
    \end{minipage}
    & 
    \begin{minipage}{\imgsizeje}
      \includegraphics[width=\linewidth]{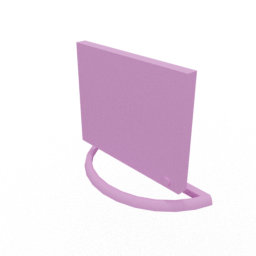}
    \end{minipage}
    & 
    \begin{minipage}{\imgsizeje}
      \includegraphics[width=\linewidth]{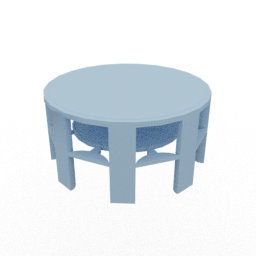}
    \end{minipage}
    & 
    \begin{minipage}{\imgsizeje}
      \includegraphics[width=\linewidth]{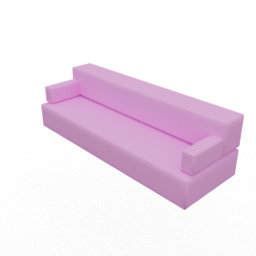}
    \end{minipage}
    & 
    \begin{minipage}{\imgsizeje}
      \includegraphics[width=\linewidth]{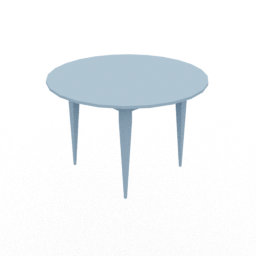}
    \end{minipage}
    & 
    \begin{minipage}{\imgsizeje}
      \includegraphics[width=\linewidth]{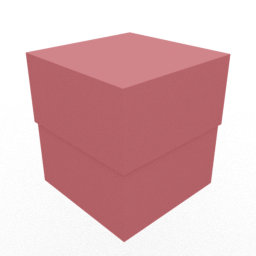}
    \end{minipage}
    \\ \hline
    \rotatebox[origin=c]{90}{3DCNN}&
    \begin{minipage}{\imgsizeje}
      \includegraphics[width=\linewidth]{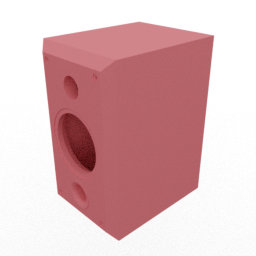}
    \end{minipage}
    &
    \begin{minipage}{\imgsizeje}
      \includegraphics[width=\linewidth]{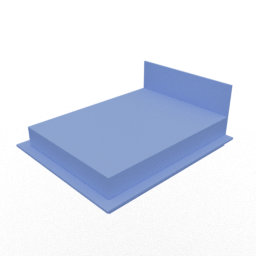}
    \end{minipage}
    & 
    \begin{minipage}{\imgsizeje}
      \includegraphics[width=\linewidth]{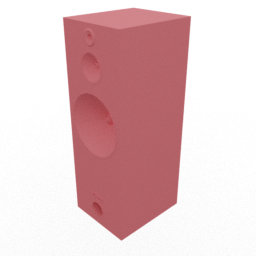}
    \end{minipage}
    & 
    \begin{minipage}{\imgsizeje}
      \includegraphics[width=\linewidth]{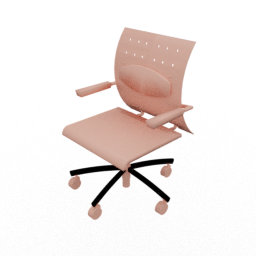}
    \end{minipage}
    & 
    \begin{minipage}{\imgsizeje}
      \includegraphics[width=\linewidth]{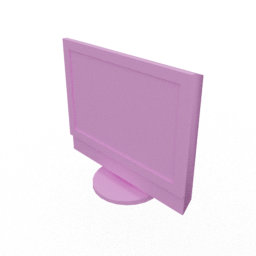}
    \end{minipage}
    & 
    \begin{minipage}{\imgsizeje}
      \includegraphics[width=\linewidth]{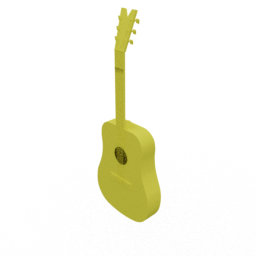}
    \end{minipage}
    & 
    \begin{minipage}{\imgsizeje}
      \includegraphics[width=\linewidth]{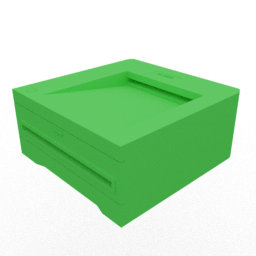}
    \end{minipage}
    & 
    \begin{minipage}{\imgsizeje}
      \includegraphics[width=\linewidth]{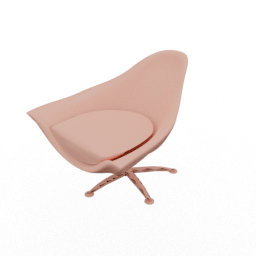}
    \end{minipage}
    & 
    \begin{minipage}{\imgsizeje}
      \includegraphics[width=\linewidth]{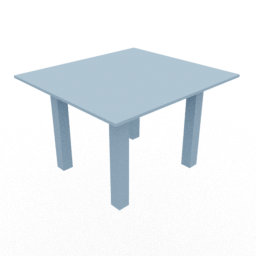}
    \end{minipage}
    & 
    \begin{minipage}{\imgsizeje}
      \includegraphics[width=\linewidth]{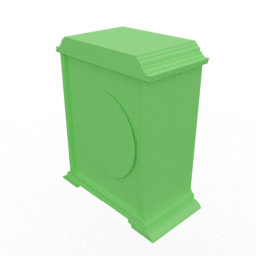}
    \end{minipage}
    \\ \hline
    \rotatebox[origin=c]{90}{JointEm.}&
    \begin{minipage}{\imgsizeje}
      \includegraphics[width=\linewidth]{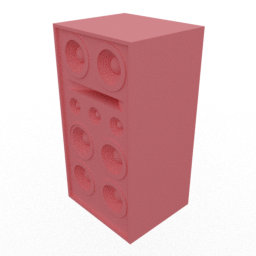}
    \end{minipage}
    &
    \begin{minipage}{\imgsizeje}
      \includegraphics[width=\linewidth]{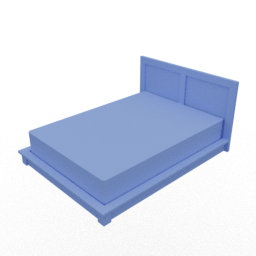}
    \end{minipage}
    & 
    \begin{minipage}{\imgsizeje}
      \includegraphics[width=\linewidth]{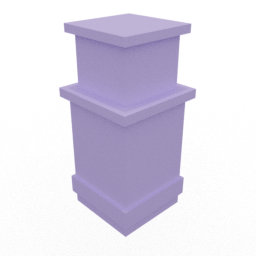}
    \end{minipage}
    & 
    \begin{minipage}{\imgsizeje}
      \includegraphics[width=\linewidth]{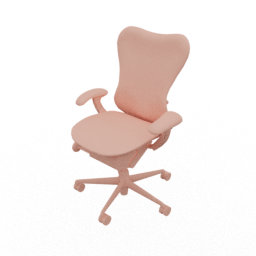}
    \end{minipage}
    & 
    \begin{minipage}{\imgsizeje}
      \includegraphics[width=\linewidth]{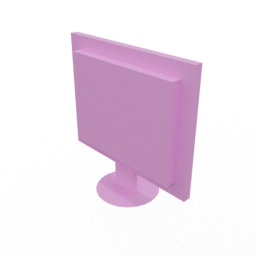}
    \end{minipage}
    & 
    \begin{minipage}{\imgsizeje}
      \includegraphics[width=\linewidth]{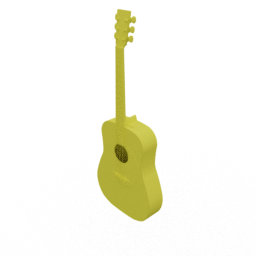}
    \end{minipage}
    & 
    \begin{minipage}{\imgsizeje}
      \includegraphics[width=\linewidth]{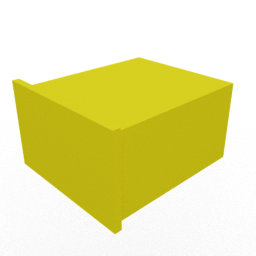}
    \end{minipage}
    & 
    \begin{minipage}{\imgsizeje}
      \includegraphics[width=\linewidth]{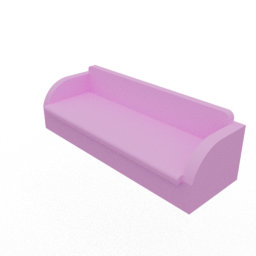}
    \end{minipage}
    & 
    \begin{minipage}{\imgsizeje}
      \includegraphics[width=\linewidth]{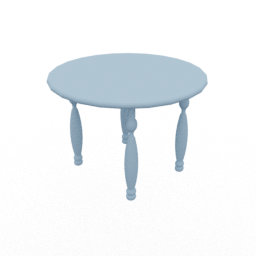}
    \end{minipage}
    & 
    \begin{minipage}{\imgsizeje}
      \includegraphics[width=\linewidth]{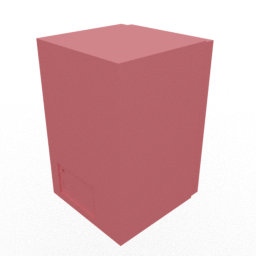}
    \end{minipage}
    \\ \hline
    \rotatebox[origin=c]{90}{\textbf{Ours}}&
    \begin{minipage}{\imgsizeje}
      \includegraphics[width=\linewidth]{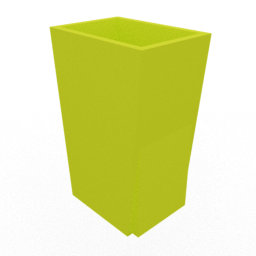}
    \end{minipage}
    &
    \begin{minipage}{\imgsizeje}
      \includegraphics[width=\linewidth]{media/Proxy Diagram/20b7fd7affe7ef07c370aa5e215a8f19.jpg}
    \end{minipage}
    & 
    \begin{minipage}{\imgsizeje}
      \includegraphics[width=\linewidth]{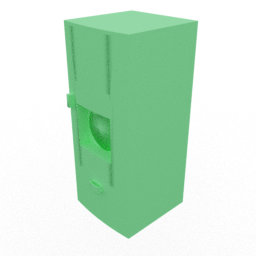}
    \end{minipage}
    & 
    \begin{minipage}{\imgsizeje}
      \includegraphics[width=\linewidth]{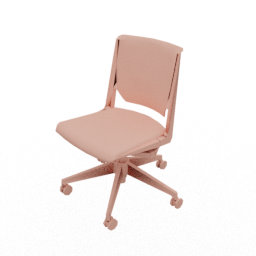}
    \end{minipage}
    & 
    \begin{minipage}{\imgsizeje}
      \includegraphics[width=\linewidth]{media/Proxy Diagram/45a4d128553abb329bf8498db368caef.jpg}
    \end{minipage}
    & 
    \begin{minipage}{\imgsizeje}
      \includegraphics[width=\linewidth]{media/Proxy Diagram/10158b6f14c2700871e863c034067817.jpg}
    \end{minipage}
    & 
    \begin{minipage}{\imgsizeje}
      \includegraphics[width=\linewidth]{media/Proxy Diagram/c2579f3b815ca02d9318709b609c3a71.jpg}
    \end{minipage}
    & 
    \begin{minipage}{\imgsizeje}
      \includegraphics[width=\linewidth]{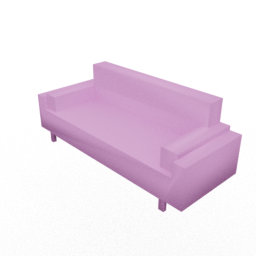}
    \end{minipage}
    & 
    \begin{minipage}{\imgsizeje}
      \includegraphics[width=\linewidth]{media/Proxy Diagram/c4e7fef0548b8d87247b7570b0fbe63d.jpg}
    \end{minipage}
    & 
    \begin{minipage}{\imgsizeje}
      \includegraphics[width=\linewidth]{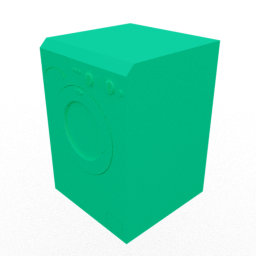}
    \end{minipage}
    \\ \hline
    \rotatebox[origin=c]{90}{GT}&
    \begin{minipage}{\imgsizeje}
      \includegraphics[width=\linewidth]{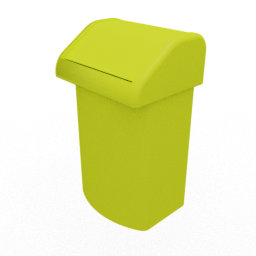}
    \end{minipage}
    &
    \begin{minipage}{\imgsizeje}
      \includegraphics[width=\linewidth]{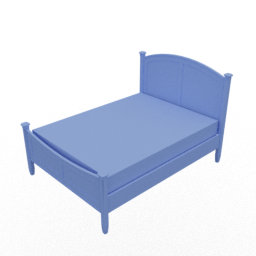}
    \end{minipage}
    & 
    \begin{minipage}{\imgsizeje}
      \includegraphics[width=\linewidth]{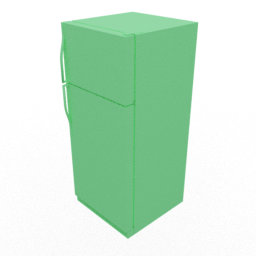}
    \end{minipage}
    & 
    \begin{minipage}{\imgsizeje}
      \includegraphics[width=\linewidth]{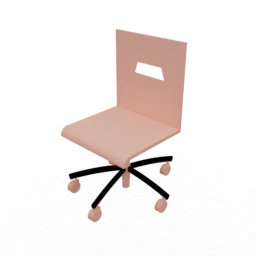}
    \end{minipage}
    & 
    \begin{minipage}{\imgsizeje}
      \includegraphics[width=\linewidth]{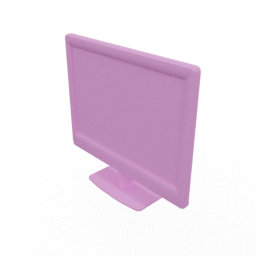}
    \end{minipage}
    & 
    \begin{minipage}{\imgsizeje}
      \includegraphics[width=\linewidth]{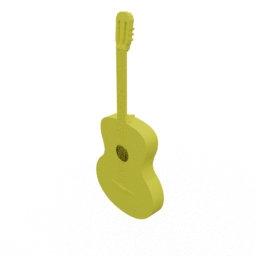}
    \end{minipage}
    & 
    \begin{minipage}{\imgsizeje}
      \includegraphics[width=\linewidth]{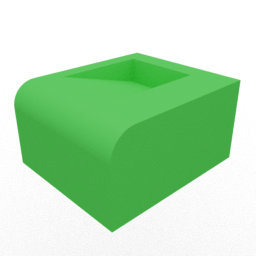}
    \end{minipage}
    & 
    \begin{minipage}{\imgsizeje}
      \includegraphics[width=\linewidth]{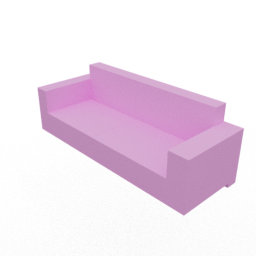}
    \end{minipage}
    & 
    \begin{minipage}{\imgsizeje}
      \includegraphics[width=\linewidth]{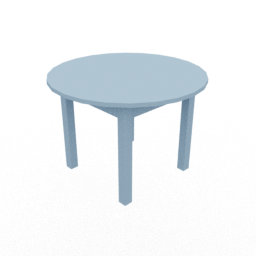}
    \end{minipage}
    & 
    \begin{minipage}{\imgsizeje}
      \includegraphics[width=\linewidth]{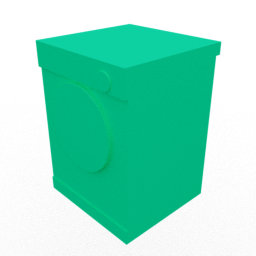}
    \end{minipage}
    \\ \hline
  \end{tabular}
  \caption{CAD retrieval for seen classes from the Scan-CAD Object Similarity Dataset~\cite{dahnert2019joint} test set.
  Compared to the hand-crafted FPFH \cite{rusu2009fast} as well as fully-supervised learned methods PointNet \cite{qi2017pointnet}, 3DCNN \cite{qi2016volumetric}, and JointEmbedding \cite{dahnert2019joint}, our weakly-supervised method is able to retrieve semantically similar models significantly more often.
}
  \label{fig:retrievals-seen}
\end{figure}

\subsubsection{Comparison to state of the art.}
We evaluate our approach on the Scan-CAD Object Similarity Dataset~\cite{dahnert2019joint}.
We compare with state-of-the-art handcrafted shape descriptors FPFH \cite{rusu2009fast} and SHOT \cite{tombari2010unique}, learned shape descriptors from PointNet \cite{qi2017pointnet} and the 3D CNN of \cite{qi2016volumetric}, and the approach of  \cite{dahnert2019joint} for CAD retrieval which constructs a triplet-driven joint embedding space.
Note that all of the learned baselines require scan-CAD association supervision, which is thus provided to them during training; our method does not use any scan-CAD association information.

For both FPFH and SHOT, we evaluate the descriptors on 1024 uniformly sampled points from the object surfaces.
PointNet~\cite{qi2017pointnet} and the 3D CNN \cite{qi2016volumetric} are trained for object classification on 1024 sampled surface points and $32^3$ occupancy grids, respectively.
We extract the feature vectors before the classification layer, yielding features of dimension 256 and 512, respectively. 
We follow the training procedure described for the JointEmbedding approach \cite{dahnert2019joint} for CAD retrieval, providing scan-CAD supervision information.
For additional baseline evaluation setup, we refer to the appendix.
Note that our method does not require any scan-CAD association information, and can be simply applied to detected object boxes in a scene.

Table~\ref{tab:comparison} evaluates the task of CAD retrieval to a scan queries for seen and unseen category evaluation scenarios from \cite{dahnert2019joint,avetisyan2019end}, respectively, using ground truth box detections.
Our end-to-end differentiable weakly-supervised approach even outperforms supervised retrieval methods across all metrics except category accuracy, which provides only a coarse reflection of retrieval performance. 
We show sample retrievals in Fig.~\ref{fig:retrievals-seen}.

We further evaluate our approach compared with state of the art given predicted oriented bounding boxes from a state-of-the-art 3D object detector \cite{you2021canonical}. 
Table~\ref{tab:metrics-predicted-bboxes} shows that all methods' performance degrade somewhat from imperfect detections; however, our approach maintains more robust performance across all evaluation metrics.

\subsubsection{Robustness to unseen class categories.}
We also evaluate CAD retrieval for scan queries of classes that were not seen during training in Table~\ref{tab:comparison} on Scan2CAD~\cite{avetisyan2019scan2cad} data, demonstrating that our approach can generalize to new categories without requiring any re-training.
In this challenging scenario, all methods' performance decrease in retrieving for unseen classes, particularly in precise retrieval measures, though our method maintains good geometric IoU approximations from retrieved objects while outperforming state of the art on all measures.
Overall, our differentiable retrieval enables significantly more robustness to unseen classes than state of the art. 
Figure~\ref{fig:retrievals-unseen} shows sample retrievals on unseen classes from Scan2CAD.
\begin{table}[t]
    \centering
    \begin{tabular}{|l|l|l|l|l|l|l|l|l|l|l|l|l|l|}
        \cline{2-13}
        \multicolumn{1}{c|}{} & \multicolumn{5}{c|}{seen classes} & & \multicolumn{6}{c|}{unseen classes} \\ \hline
        Method                         & Top1 & Cat & IoU$_1$ & IoU$_5$ & RQ & & Top1 & Top5 & Cat & IoU$_1$ & IoU$_5$ & MRR \\ \hline \hline
        Random                         & 0.02 & 0.04 & 0.13 & 0.12 & 0.01 & & 0.00 & 0.00 & 0.01 & 0.13 & 0.14 & 0.00\\ 
        SHOT \cite{tombari2010unique}  & 0.02 & 0.04 & 0.12 & 0.13 & 0.01 & & 0.00 & 0.00 & 0.01 & 0.13 & 0.13 & 0.00\\ 
        FPFH \cite{rusu2009fast}       & 0.07 & 0.13 & 0.15 & 0.16 & 0.03 & & 0.00 & 0.00 & 0.03 & 0.13 & 0.13 & 0.00\\
        3DCNN (class.)* \cite{qi2016volumetric} & 0.35 & 0.58 & 0.45 & 0.48 & 0.17 & & 0.04 & 0.09 & 0.39 & 0.32 & 0.32 & 0.07 \\  
        PointNet* \cite{qi2017pointnet}& 0.35 & 0.60 & 0.48 & 0.46 & 0.17 & & 0.03 & 0.11 & 0.47 & 0.32 & 0.33 & 0.09\\ 
        JointEmbedding* \cite{dahnert2019joint} & 0.43 & \bf{0.68} & - & - & 0.18 & & - & - & - & - & - & -   \\
        JointEmbedding*$\dagger$       & 0.42 & 0.62 & 0.50 & 0.47 & 0.19 & & 0.02 & 0.05 & 0.15 & 0.31 & 0.31 & 0.03\\
        \hline
        Ours &  \bf{0.48} & 0.66 & \bf{0.54} & \bf{0.53} & \bf{0.23} & & \textbf{0.11} & \textbf{0.28} & \textbf{0.57} & \textbf{0.46} & \textbf{0.43} & \textbf{0.19}\\ \hline
    \end{tabular}
    \caption{CAD retrieval to scan objects in seen and unseen category evaluation settings using \cite{dahnert2019joint} and \cite{avetisyan2019scan2cad}, respectively. Our differentiable retrieval achieves comparable and even improved performance without using any scan-CAD association annotations. Methods with a * require scan-CAD association supervision. $\dagger$ indicates reproduced results. }
    \label{tab:comparison}
\end{table}
\newcommand\imgsizescene{.24\textwidth}
\begin{figure}[t]
  \centering
  \begin{tabular}{| c | c | c | c |}
    \hline
    Input Scan & JointEmbedding & Ours & GT 
    \\ \hline
    \begin{minipage}{\imgsizescene}
      \centerline{\includegraphics[width=1.03\linewidth]{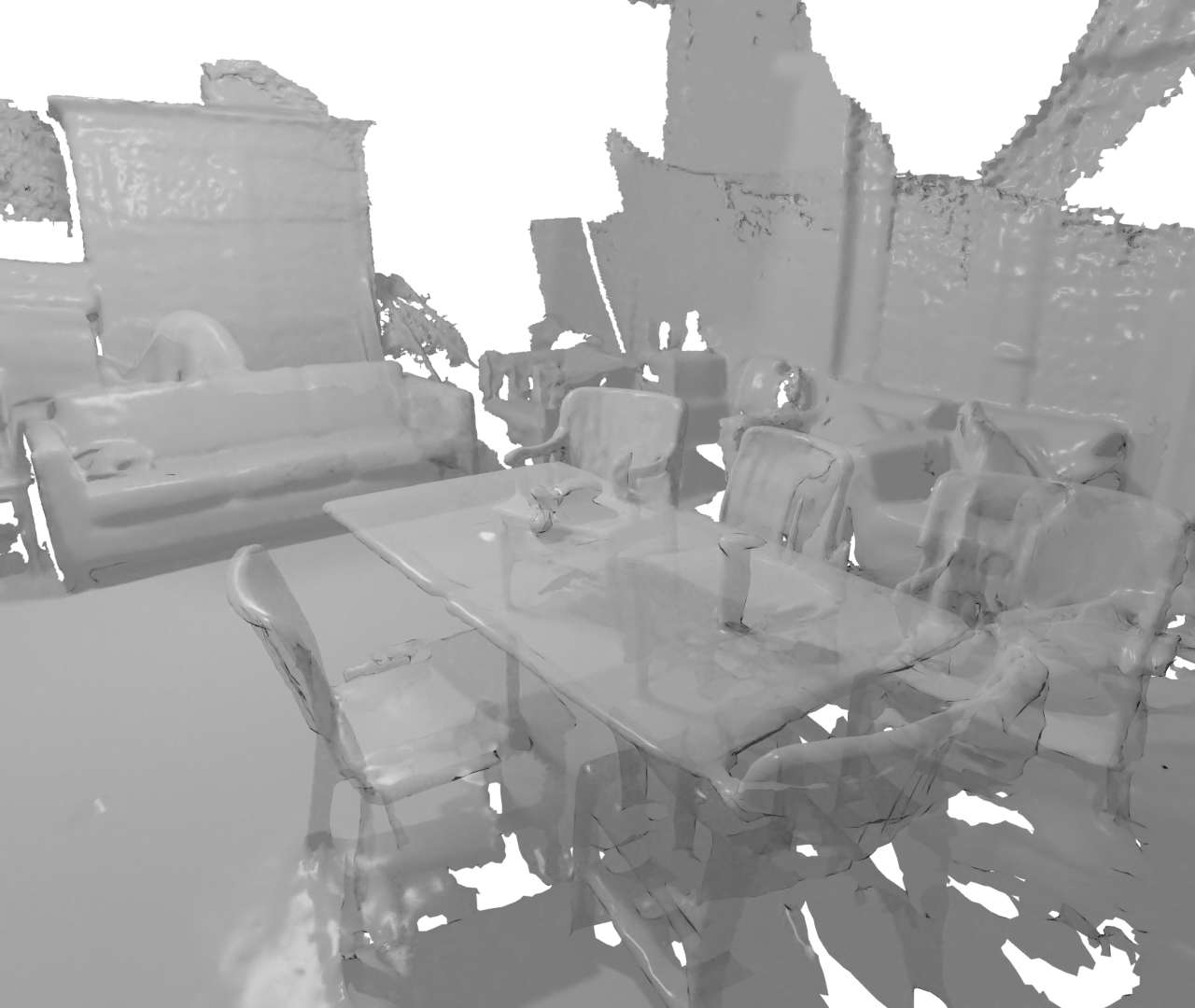}}
    \end{minipage}
    &
    \begin{minipage}{\imgsizescene}
      \centerline{\includegraphics[width=1.03\linewidth]{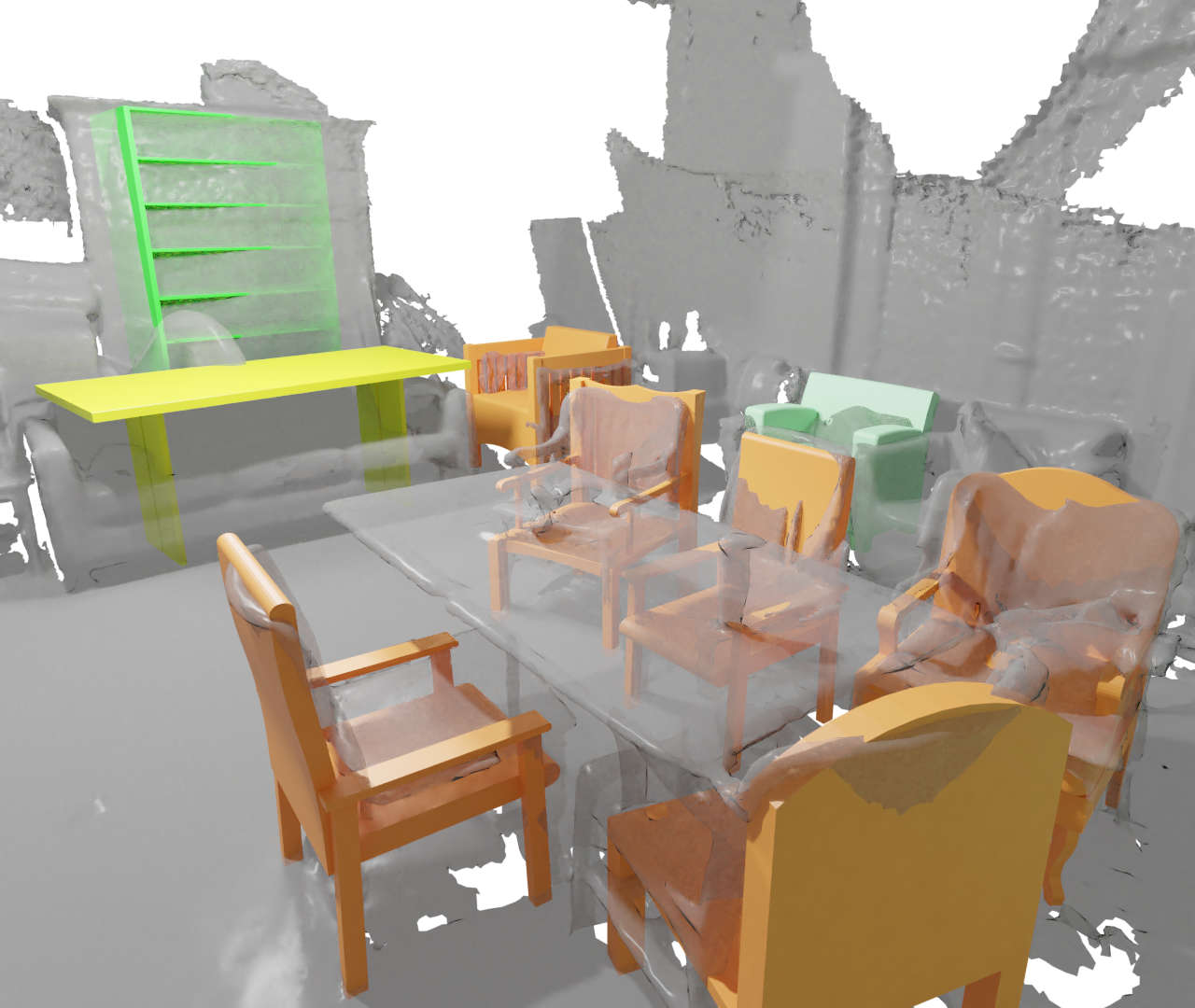}}
    \end{minipage}
    &
    \begin{minipage}{\imgsizescene}
      \centerline{\includegraphics[width=1.03\linewidth]{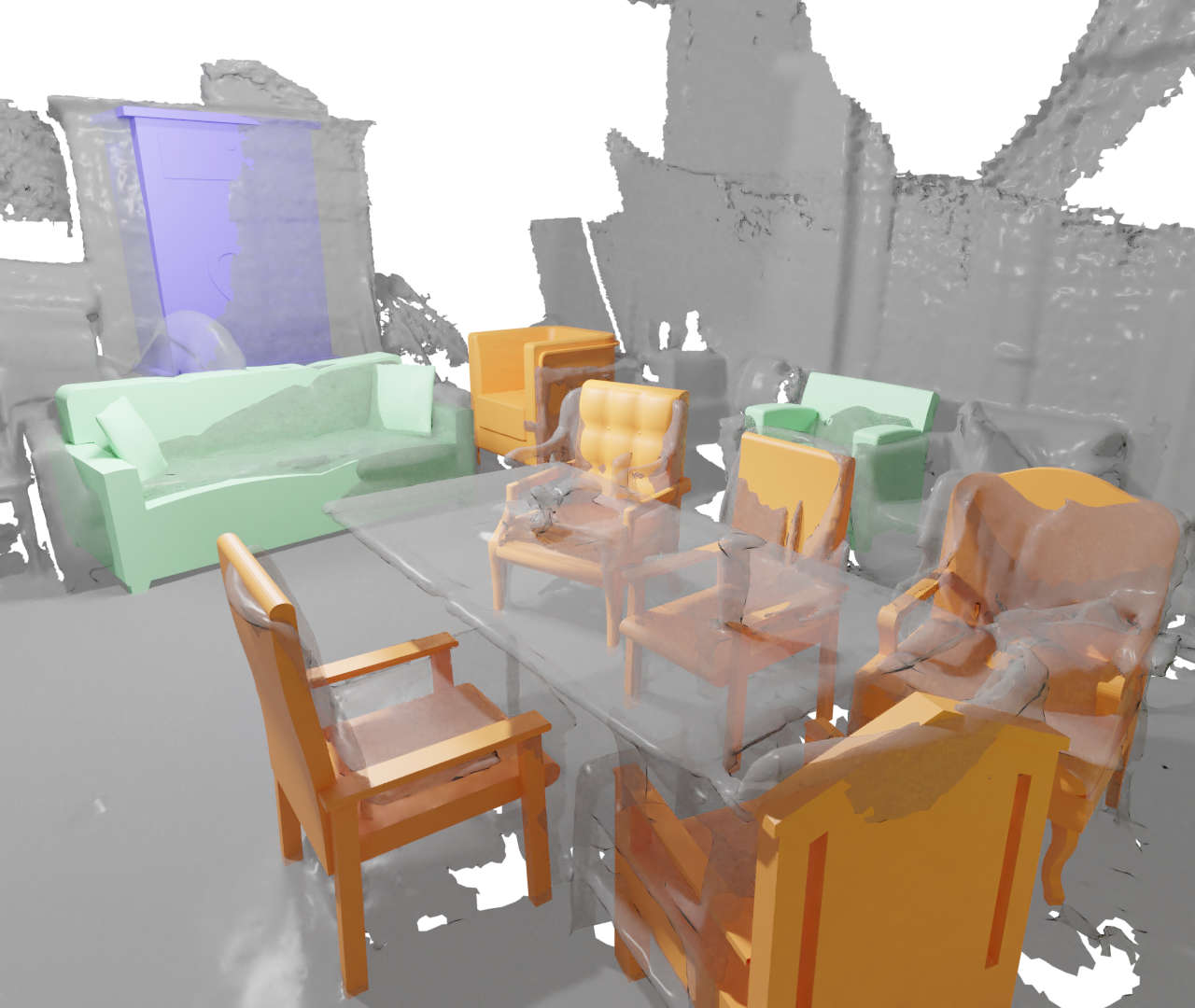}}
    \end{minipage}
    &
    \begin{minipage}{\imgsizescene}
      \centerline{\includegraphics[width=1.03\linewidth]{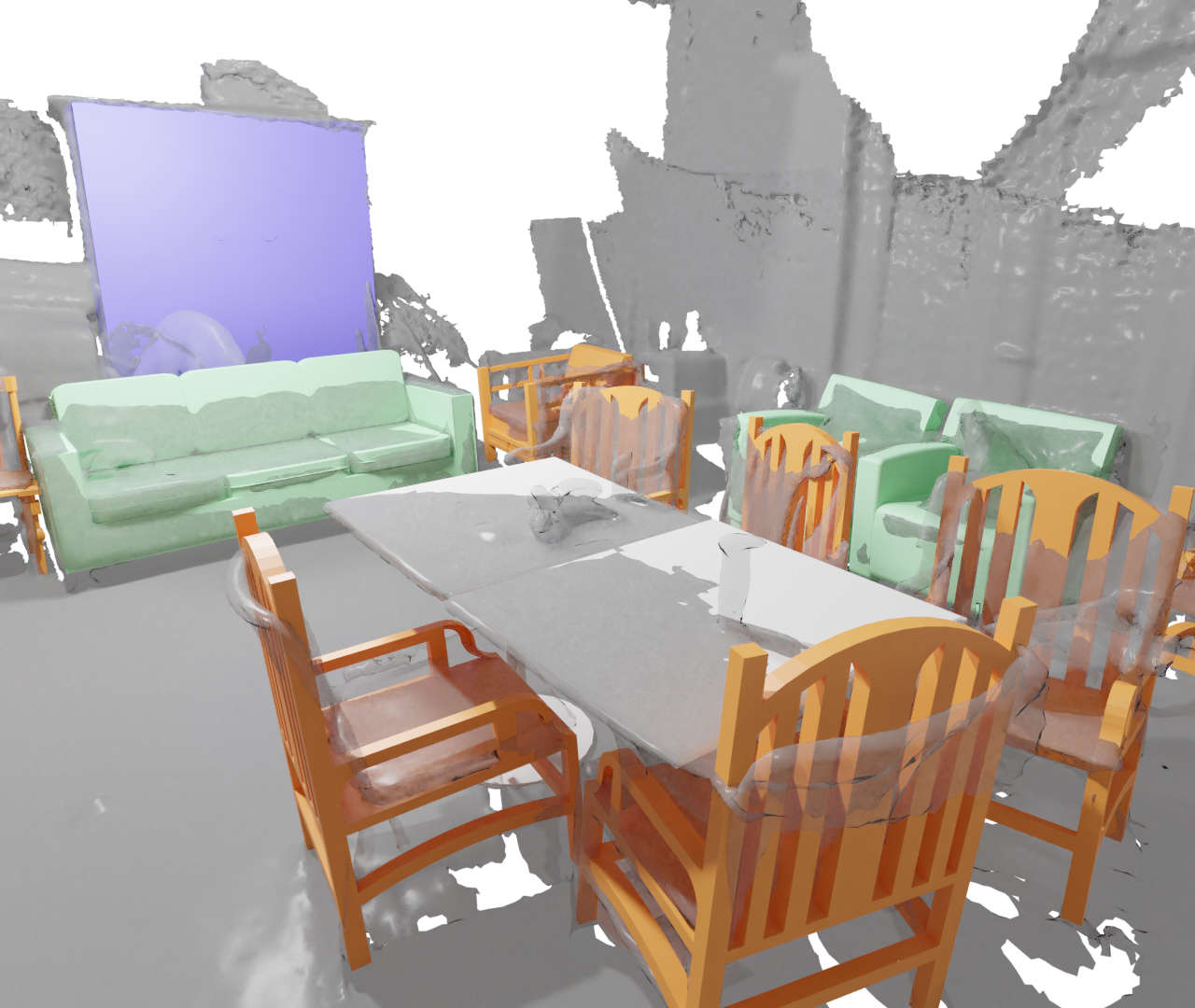}}
    \end{minipage}
    
    \\ \hline 
    \begin{minipage}{\imgsizescene}
      \centerline{\includegraphics[width=1.03\linewidth]{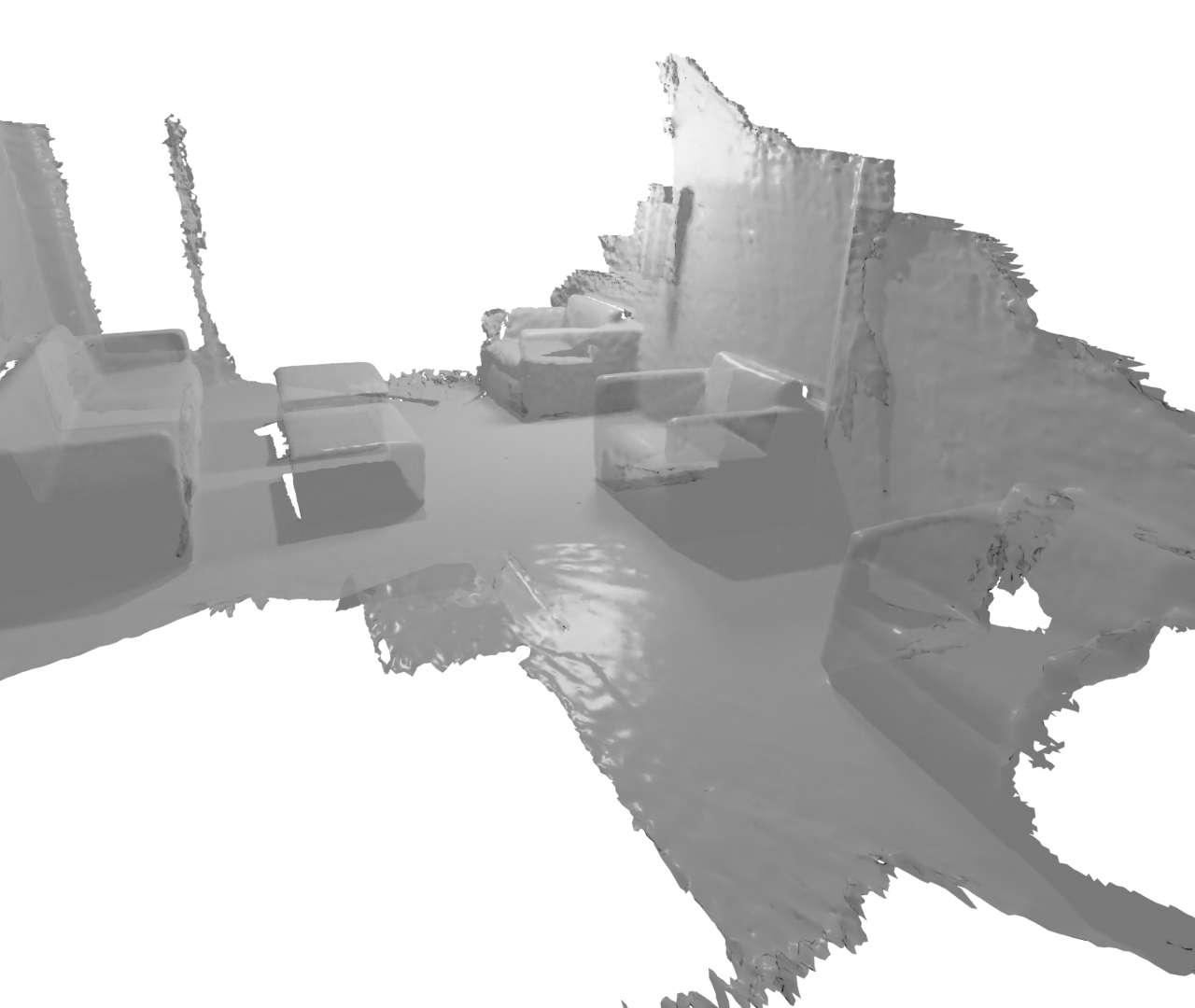}}
    \end{minipage}
    &
    \begin{minipage}{\imgsizescene}
      \centerline{\includegraphics[width=1.03\linewidth]{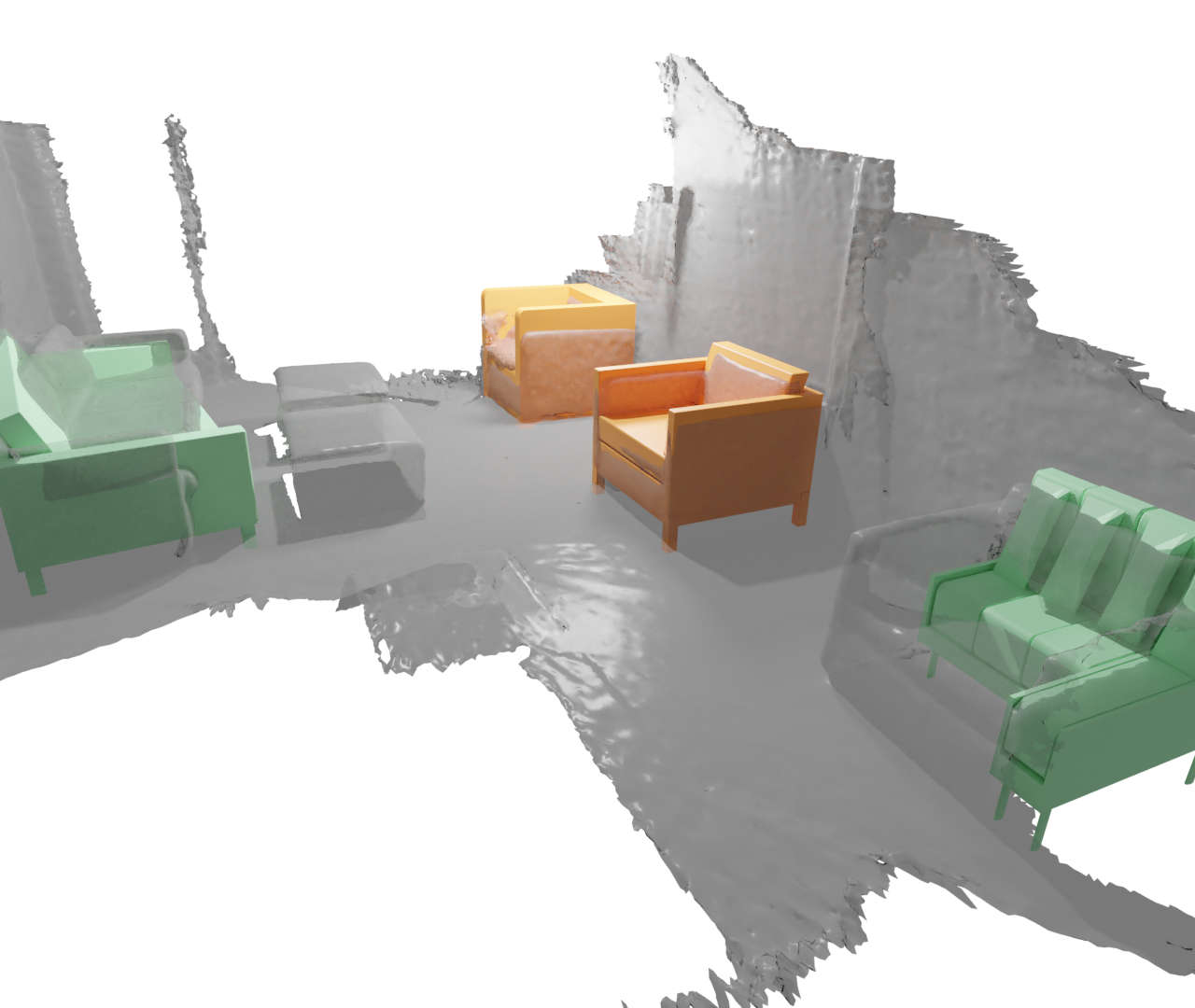}}
    \end{minipage}
    & 
    \begin{minipage}{\imgsizescene}
      \centerline{\includegraphics[width=1.03\linewidth]{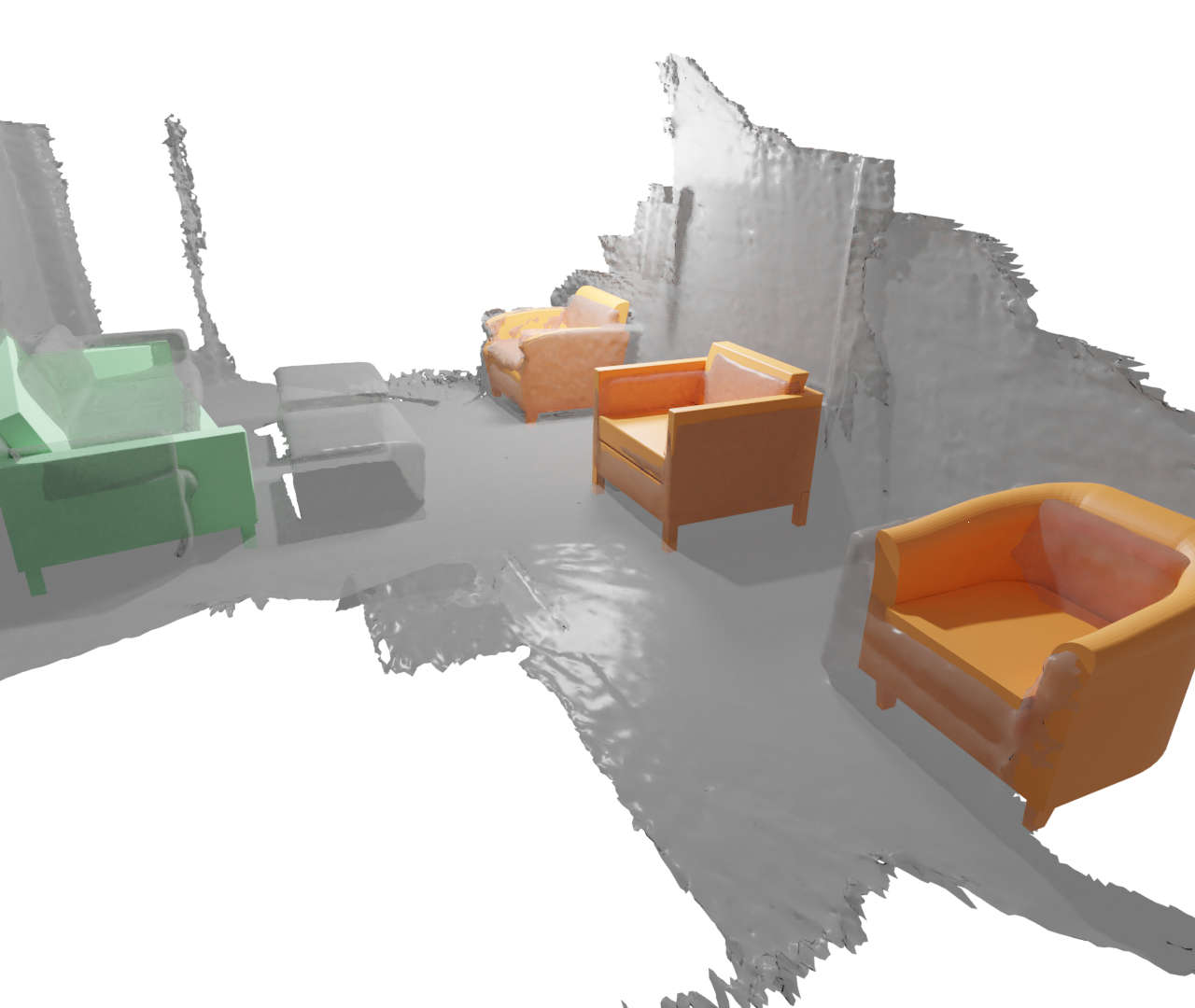}}
    \end{minipage}
    &
    \begin{minipage}{\imgsizescene}
      \centerline{\includegraphics[width=1.03\linewidth]{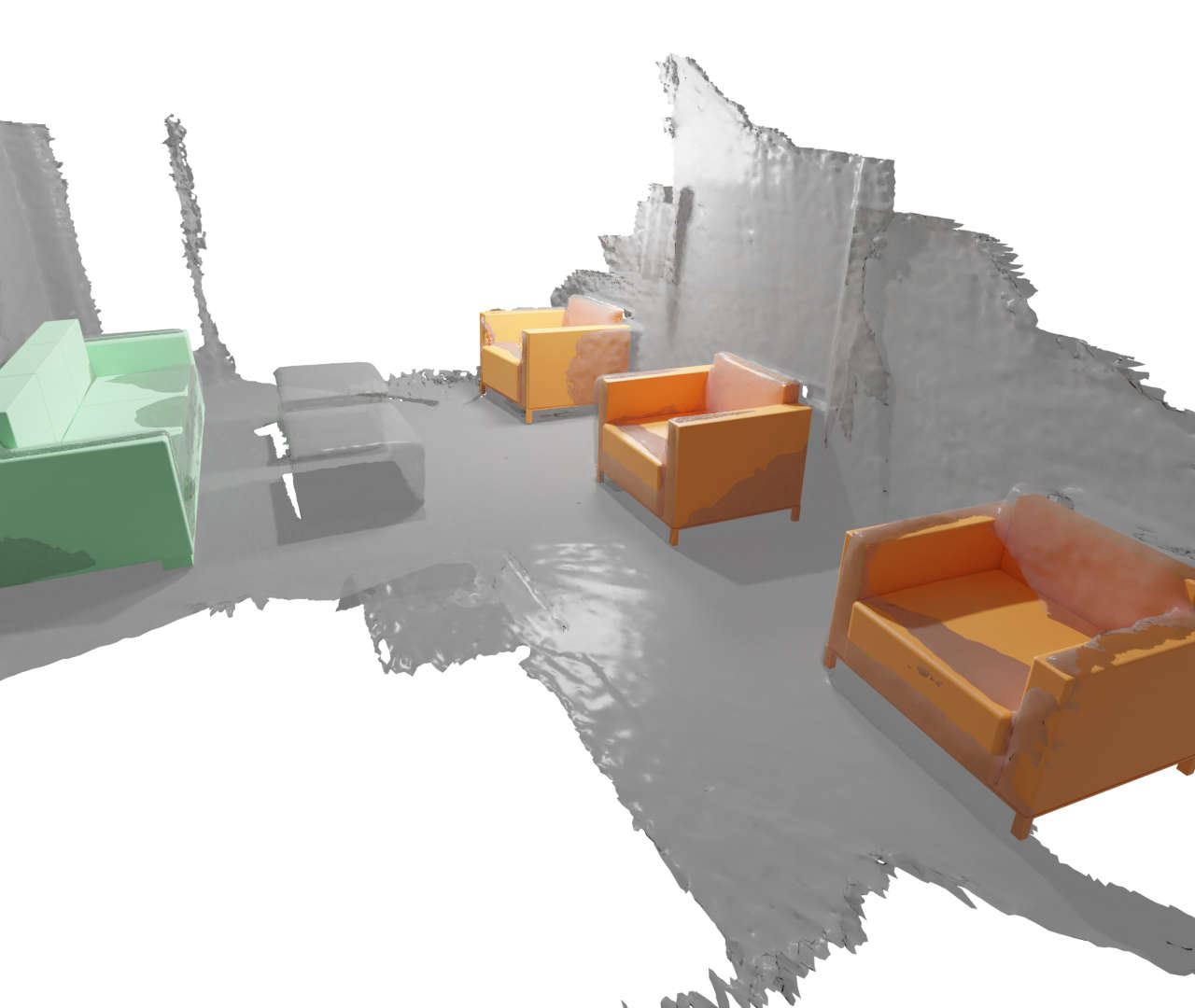}}
    \end{minipage}
    \\ \hline
  \end{tabular}
  \caption{Qualitative illustration of full-scene reconstructions using our weakly-supervised method for CAD retrieval to scan objects detected with predicted OBBs from CanonicalVoting~\cite{you2021canonical} on ScanNet data.
  Our weakly-supervsed approach can outperform the fully-supervised JointEmbedding~\cite{dahnert2019joint} for CAD retrieval, with more representative CAD models retrieved for the sofa and cabinet (top) as well as chair (bottom).
}
  \label{fig:full-reconstructions}
\end{figure}

\begin{table}[t]
    \centering
    \begin{tabular}{|l|l|l|l|l|l|l|l|l|l|l|l|l|l|}
        \cline{2-13}
        \multicolumn{1}{c|}{} & \multicolumn{5}{c|}{seen classes} & & \multicolumn{6}{c|}{unseen classes} \\ \hline 
        Method         & Top1 & Cat & IoU$_1$ & IoU$_5$ & RQ & & Top1 & Top5 & Cat & IoU$_1$ & IoU$_5$ & MRR \\
        \hline \hline
        3DCNN* \cite{qi2016volumetric} & 0.26 & 0.48 & 0.37 & 0.37 & 0.13 & & 0.00 & 0.01 & 0.09 & 0.22 & 0.21 & 0.01\\
        PointNet* \cite{qi2017pointnet}& 0.15 & 0.29 & 0.31 & 0.24 & 0.07 & & 0.01 & 0.03 & 0.20 & 0.21 & 0.21 & 0.02\\
        JointEmbedding* \cite{dahnert2019joint} & 0.13 & 0.25 & 0.26 & 0.24 & 0.06 & & 0.01 & 0.04 & 0.31 & 0.29 & 0.28 & 0.04\\   
        \hline                         
        Ours & \textbf{0.35} & \textbf{0.56} & \textbf{0.46} & \textbf{0.45} & \textbf{0.17} & & \textbf{0.05} & \textbf{0.12} & \textbf{0.43} & \textbf{0.39} & \textbf{0.38} & \textbf{0.09}\\ \hline
    \end{tabular}                                                                        
    \caption{CAD retrieval with predicted oriented bounding boxes from CanonicalVoting~\cite{you2021canonical}.
    Our weakly-supervised approach maintains improved performance in comparison with state-of-the-art learned methods requiring full scan-CAD association supervision (denoted by *).
    }
    \label{tab:metrics-predicted-bboxes}
\end{table}

\newcommand\imgsize{.128\textwidth}

\begin{figure}[!t]
  \centering
  \begin{tabular}{ | c | c | c | c | c | c | c | c |}
    \hline
    \rotatebox[origin=c]{90}{Scan Query}
    & 
    \begin{minipage}{\imgsize}
      \includegraphics[width=\linewidth]{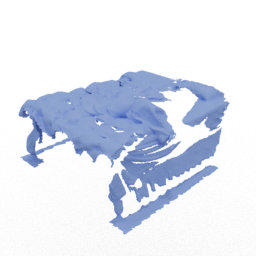}
    \end{minipage}
    &
    \begin{minipage}{\imgsize}
      \includegraphics[width=\linewidth]{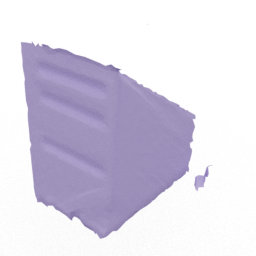}
    \end{minipage}
    &
    \begin{minipage}{\imgsize}
      \includegraphics[width=\linewidth]{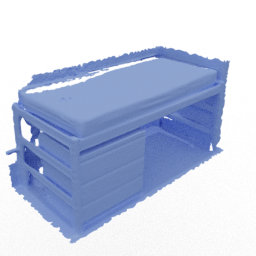}
    \end{minipage}
    & 
    \begin{minipage}{\imgsize}
      \includegraphics[width=\linewidth]{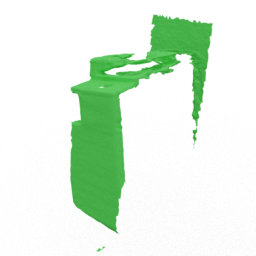}
    \end{minipage}
    & 
    \begin{minipage}{\imgsize}
      \includegraphics[width=\linewidth]{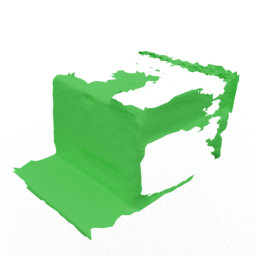}
    \end{minipage}
    & 
    \begin{minipage}{\imgsize}
      \includegraphics[width=\linewidth]{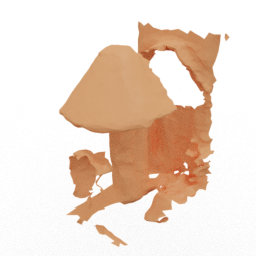}
    \end{minipage}
    & 
    \begin{minipage}{\imgsize}
      \includegraphics[width=\linewidth]{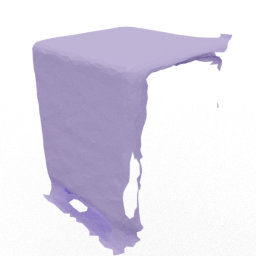}
    \end{minipage}
    \\ \hline
    \rotatebox[origin=c]{90}{FPFH}&
    \begin{minipage}{\imgsize}
      \includegraphics[width=\linewidth]{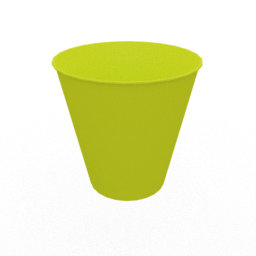}
    \end{minipage}
    &
    \begin{minipage}{\imgsize}
      \includegraphics[width=\linewidth]{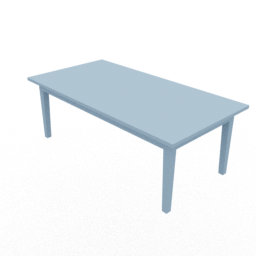}
    \end{minipage}
    & 
    \begin{minipage}{\imgsize}
      \includegraphics[width=\linewidth]{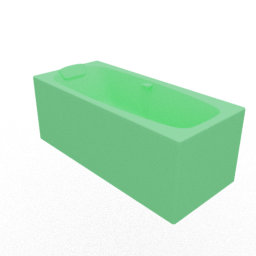}
    \end{minipage}
    & 
    \begin{minipage}{\imgsize}
      \includegraphics[width=\linewidth]{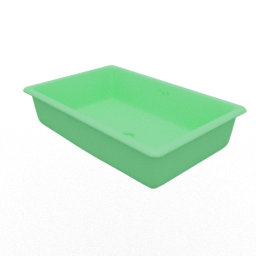}
    \end{minipage} 
    & 
    \begin{minipage}{\imgsize}
      \includegraphics[width=\linewidth]{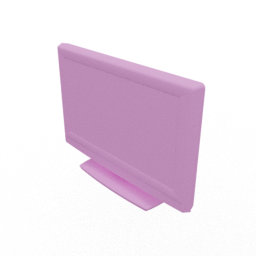}
    \end{minipage}
    & 
    \begin{minipage}{\imgsize}
      \includegraphics[width=\linewidth]{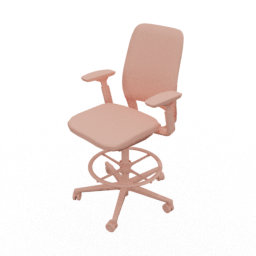}
    \end{minipage}
    & 
    \begin{minipage}{\imgsize}
      \includegraphics[width=\linewidth]{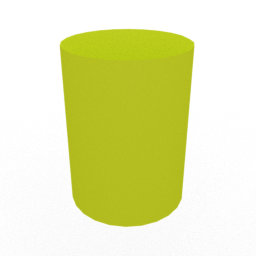}
    \end{minipage}
    \\ \hline  
    \rotatebox[origin=c]{90}{PointNet}&
    \begin{minipage}{\imgsize}
      \includegraphics[width=\linewidth]{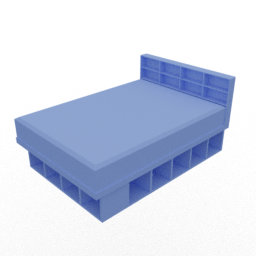}
    \end{minipage}
    &
    \begin{minipage}{\imgsize}
      \includegraphics[width=\linewidth]{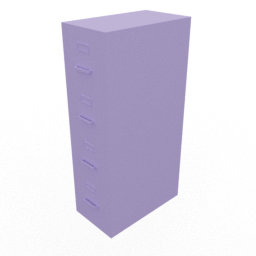}
    \end{minipage}
    & 
    \begin{minipage}{\imgsize}
      \includegraphics[width=\linewidth]{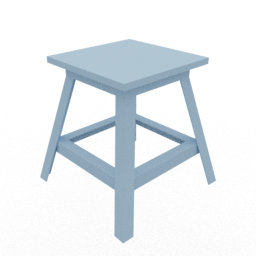}
    \end{minipage}
    & 
    \begin{minipage}{\imgsize}
      \includegraphics[width=\linewidth]{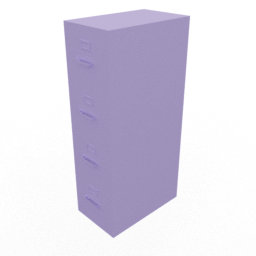}
    \end{minipage}
    & 
    \begin{minipage}{\imgsize}
      \includegraphics[width=\linewidth]{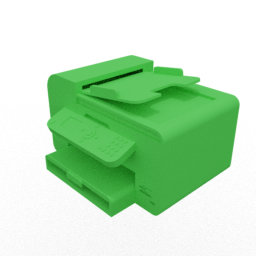}
    \end{minipage}
    & 
    \begin{minipage}{\imgsize}
      \includegraphics[width=\linewidth]{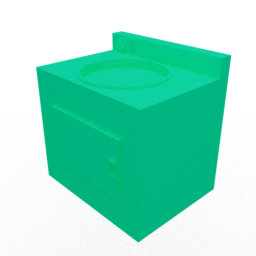}
    \end{minipage}
    & 
    \begin{minipage}{\imgsize}
      \includegraphics[width=\linewidth]{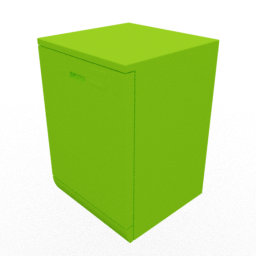}
    \end{minipage}
    \\ \hline
    \rotatebox[origin=c]{90}{3DCNN}&
    \begin{minipage}{\imgsize}
      \includegraphics[width=\linewidth]{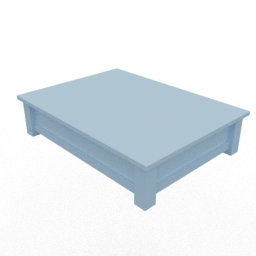}
    \end{minipage}
    &
    \begin{minipage}{\imgsize}
      \includegraphics[width=\linewidth]{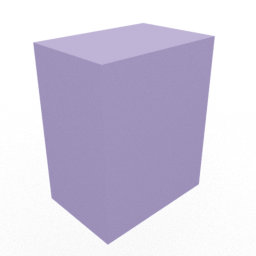}
    \end{minipage}
    & 
    \begin{minipage}{\imgsize}
      \includegraphics[width=\linewidth]{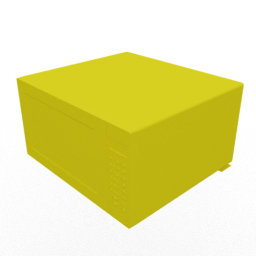}
    \end{minipage}
    & 
    \begin{minipage}{\imgsize}
      \includegraphics[width=\linewidth]{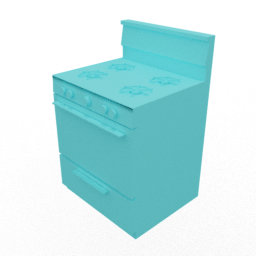}
    \end{minipage}
    & 
    \begin{minipage}{\imgsize}
      \includegraphics[width=\linewidth]{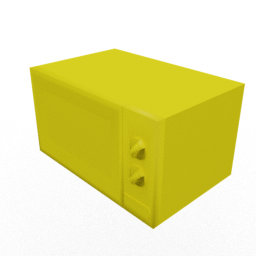}
    \end{minipage}
    & 
    \begin{minipage}{\imgsize}
      \includegraphics[width=\linewidth]{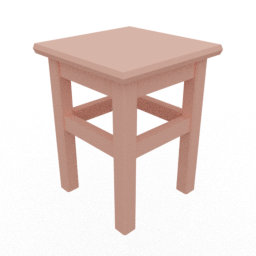}
    \end{minipage}
    & 
    \begin{minipage}{\imgsize}
      \includegraphics[width=\linewidth]{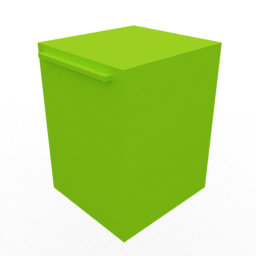}
    \end{minipage}
    \\ \hline
    \rotatebox[origin=c]{90}{JointEmb.}&
    \begin{minipage}{\imgsize}
      \includegraphics[width=\linewidth]{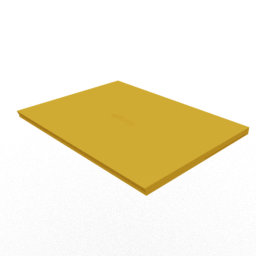}
    \end{minipage}
    &
    \begin{minipage}{\imgsize}
      \includegraphics[width=\linewidth]{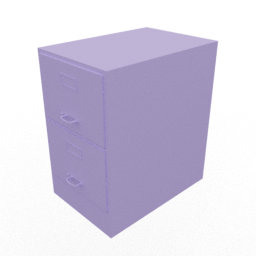}
    \end{minipage}
    & 
    \begin{minipage}{\imgsize}
      \includegraphics[width=\linewidth]{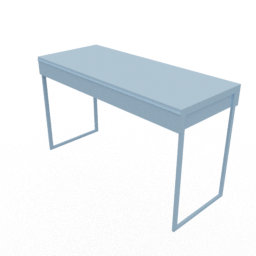}
    \end{minipage}
    & 
    \begin{minipage}{\imgsize}
      \includegraphics[width=\linewidth]{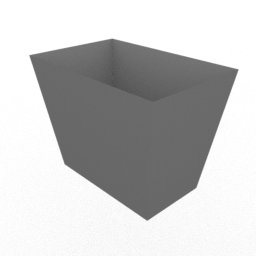}
    \end{minipage}
    & 
    \begin{minipage}{\imgsize}
      \includegraphics[width=\linewidth]{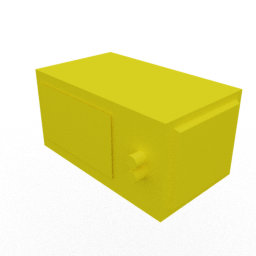}
    \end{minipage}
    & 
    \begin{minipage}{\imgsize}
      \includegraphics[width=\linewidth]{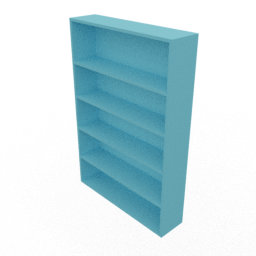}
    \end{minipage}
    & 
    \begin{minipage}{\imgsize}
      \includegraphics[width=\linewidth]{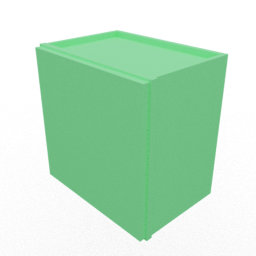}
    \end{minipage}
    \\ \hline
    \rotatebox[origin=c]{90}{Ours}&
    \begin{minipage}{\imgsize}
      \includegraphics[width=\linewidth]{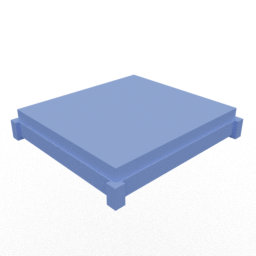}
    \end{minipage}
    &
    \begin{minipage}{\imgsize}
      \includegraphics[width=\linewidth]{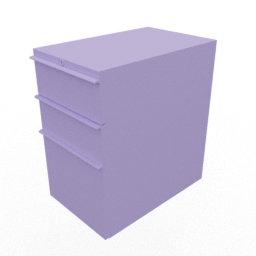}
    \end{minipage}
    & 
    \begin{minipage}{\imgsize}
      \includegraphics[width=\linewidth]{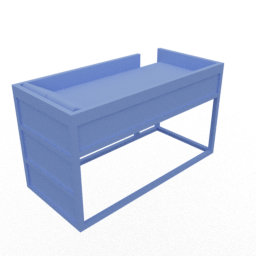}
    \end{minipage}
    & 
    \begin{minipage}{\imgsize}
      \includegraphics[width=\linewidth]{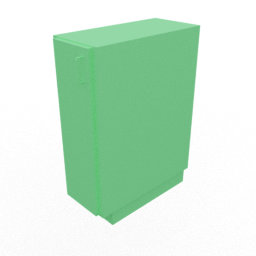}
    \end{minipage}
    & 
    \begin{minipage}{\imgsize}
      \includegraphics[width=\linewidth]{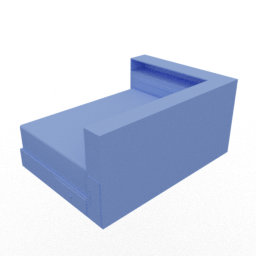}
    \end{minipage}
    & 
    \begin{minipage}{\imgsize}
      \includegraphics[width=\linewidth]{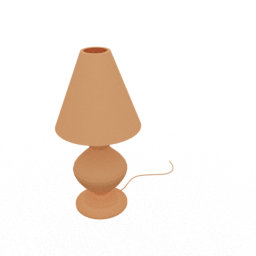}
    \end{minipage}
    & 
    \begin{minipage}{\imgsize}
      \includegraphics[width=\linewidth]{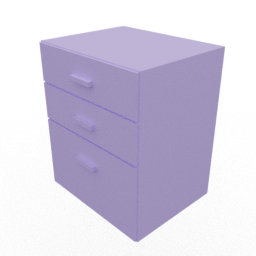}
    \end{minipage}
    \\ \hline
    \rotatebox[origin=c]{90}{GT Annotation}&
    \begin{minipage}{\imgsize}
      \includegraphics[width=\linewidth]{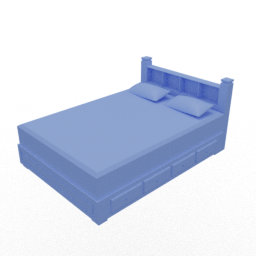}
    \end{minipage}
    &
    \begin{minipage}{\imgsize}
      \includegraphics[width=\linewidth]{media/Proxy Diagram/bfcdaac4627b7c13a7a7a90dc2dc5bd.jpg}
    \end{minipage}
    & 
    \begin{minipage}{\imgsize}
      \includegraphics[width=\linewidth]{media/Proxy Diagram/64c2347dfd79c63a63d977b06bbd429d.jpg}
    \end{minipage}
    & 
    \begin{minipage}{\imgsize}
      \includegraphics[width=\linewidth]{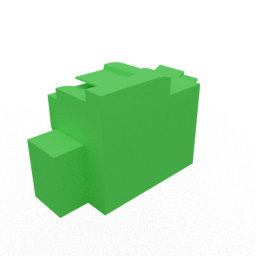}
    \end{minipage}
    & 
    \begin{minipage}{\imgsize}
      \includegraphics[width=\linewidth]{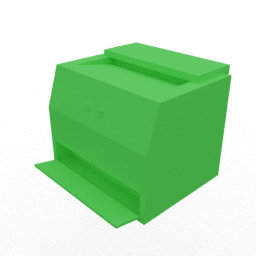}
    \end{minipage}
    & 
    \begin{minipage}{\imgsize}
      \includegraphics[width=\linewidth]{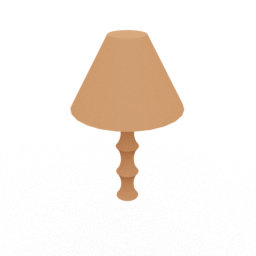}
    \end{minipage}
    & 
    \begin{minipage}{\imgsize}
      \includegraphics[width=\linewidth]{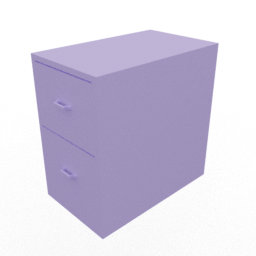}
    \end{minipage}
    \\ \hline
  \end{tabular}
  \caption{CAD retrieval shown for randomly selected scans of unseen classes from Scan2CAD~\cite{avetisyan2019scan2cad}. In comparison with state of the art, our weakly-supervised approach retrieves the correct object class and instance much more reliably than the hand-crafted features of FPFH~\cite{rusu2009fast}, as well as the fully-supervised learned methods PointNet~\cite{qi2017pointnet}, 3DCNN~\cite{qi2016volumetric}, and JointEmbedding~\cite{dahnert2019joint}.
  }
  \label{fig:retrievals-unseen}
\end{figure}

\subsubsection{Embedding space construction based on estimated soft similarities.}
Our approach guides the scan-CAD embedding space construction based on estimated geometric and perceptual similarity measure between sampled scanned objects and CADs.
Table~\ref{tab:ablation-seen} shows that this approach improves notably over the alternative of using estimated positives and negatives with a triplet loss as well as the alternative of a direct loss on the embedding space.
For the triplet loss, we employ positive and negative association based on the highest similarity CAD being a positive and the rest negative, which we empirically found to perform the best among a variety of positive/negative mining strategies (see the appendix for more detail).
The embedding loss uses $\text{MSE}(\mathbf{S}, \mathbf{P})$ to encourage the predicted cosine similarity of a scan and CAD embedding $\mathbf{S}$ to be close to the ``real" similarity as judged by the proxy metric $\mathbf{P}$.
Our soft-similarity-based embedding construction outperforms these alternatives across both geometric and retrieval metrics.

\subsubsection{What is the impact of differentiable retrieval?}
We additionally compare our embedding learning in the embedding space with an end-to-end differentiable approach using differentiable top-$k$ in lines 2 and 3 of Table~\ref{tab:ablation-seen}.
We use the loss formulation specified in Section~\ref{sec:losses} and keep the underlying similarity metric the same.
We see significant improvements across all metrics on unseen classes. 
On known classes, performance remains approximately the same.

\subsubsection{What is the impact of differentiable top-1 vs. top-$k$?}
We also investigate the impact of the number of $k$ retrievals from the top-$k$ layer during training.
Table~\ref{tab:ablation-seen} shows that with larger values of $k$ ($k=5$) we achieve improved results; for higher values ($k>5$) we found diminishing gains.

\subsubsection{How much does proxy metric selection matter?}
We estimate proxy similarity based on geometric and perceptual measures, and analyze the effect of each in the last three rows of Table~\ref{tab:ablation-seen}.
Both signals are complementary, providing the best performance together, but with more information given by the perceptual proxy metric.
We refer to the appendix for a detailed analysis of retrieval by directly using proxy metrics for retrieval; under this scenario we found that our approach maintains a performance improvement but importantly, the computational cost across a large CAD database is enormous (e.g., the perceptual comparison alone requires a comparison of $5\times8\times8\times512=163,840$ dimensions).
In contrast, for the given database of 3049 CAD objects, a query takes about one second using the combined proxy metric, while our method can embed and retrieve 250 objects per second on a Nvidia GTX 1080 Ti, which is more than two orders of magnitude faster.

\begin{table}[!ht]
    \centering
    \begin{tabular}{|l|l|l|l|l|l|l|}
    \hline
        Method & Proxy Metric & Top1 & Cat & IoU & IoU (Top 5) & RQ \\ \hline
        \hline
        Triplet Loss & perceptual & 0.34 & 0.56 & 0.47 & 0.45 & 0.17 \\
        Embedding MSE & perceptual & 0.43 & 0.62 & 0.51 & 0.48 & 0.21 \\
        Top-K training (k=1) & perceptual & 0.44 & 0.58 & 0.51 & 0.50 & 0.22 \\
        Top-K training (k=5) & geometric & 0.40 & 0.59 & 0.50 & 0.48 & 0.19 \\
        Top-K training (k=5) & perceptual & 0.46 & 0.63 & 0.52 & 0.51 & 0.22 \\
        Ours, Top-K training (k=5) & combined & \textbf{0.48} & \textbf{0.66} & \textbf{0.54} & \textbf{0.53} & \textbf{0.23} \\
    \hline
    \end{tabular}
    \caption{Ablation study on the Scan-CAD Object Similarity Dataset~\cite{dahnert2019joint} (seen classes).}
    \label{tab:ablation-seen}
\end{table}
\paragraph{Limitations.}
While our method provides a promising step towards mapping together real scans and synthetic CAD objects without full supervision, several notable limitations remain.
For instance, our method relies on annotations for oriented bounding boxes of objects in the scanned scenes, whereas a differentiable alignment optimization incorporating ideas from fully-supervised alignment methods \cite{avetisyan2019end,gumeli2021roca} could potentially allow for both retrieval and alignment estimation in the absence of any supervision.
Additionally, similarities in fine detail are not effectively captured at coarse voxel resolutions, where sparse or implicit representations could additionally inform finer-scale deformations following CAD retrieval to better fit to real-world scan observations.
Further work could make use of the RGB-color information that is recorded by most scanning devices.   
Finally, our method is limited to the use of a fixed set of CAD models. 
A more general approach could employ parametric deformation of CAD models to better fit the scan objects.

\section{Conclusion}
We have presented a weakly-supervised approach that learns CAD model retrieval to noisy and incomplete real-world scans without requiring any expensive 1:1 CAD model to scan object annotations.
Our approach to estimate perceptual- and geometrically-guided differentiable top-$k$ retrieval learns robust CAD model retrieval in an end-to-end fashion, outperforming fully-supervised state of the art methods.
We demonstrate notable improvements for seen and in particular for unseen class categories.
We believe this provides a first step towards CAD-based 3D perception in the wild.

\section*{Acknowledgements}
\vspace{-0.3cm}
This project is funded by the Bavarian State Ministry of Science and the Arts and coordinated by the Bavarian Research Institute for Digital Transformation (bidt).

%\clearpage
\bibliographystyle{splncs04}
\bibliography{egbib}

\begin{thebibliography}{10}
\providecommand{\url}[1]{\texttt{#1}}
\providecommand{\urlprefix}{URL }
\providecommand{\doi}[1]{https://doi.org/#1}

\bibitem{avetisyan2019scan2cad}
Avetisyan, A., Dahnert, M., Dai, A., Savva, M., Chang, A.X., Nie{\ss}ner, M.:
  Scan2cad: Learning cad model alignment in rgb-d scans. In: Proceedings of the
  IEEE/CVF Conference on Computer Vision and Pattern Recognition. pp.
  2614--2623 (2019)

\bibitem{avetisyan2019end}
Avetisyan, A., Dai, A., Nie{\ss}ner, M.: End-to-end cad model retrieval and
  9dof alignment in 3d scans. In: Proceedings of the IEEE/CVF International
  Conference on Computer Vision. pp. 2551--2560 (2019)

\bibitem{berthet2020learning}
Berthet, Q., Blondel, M., Teboul, O., Cuturi, M., Vert, J.P., Bach, F.:
  Learning with differentiable pertubed optimizers. Advances in neural
  information processing systems  \textbf{33},  9508--9519 (2020)

\bibitem{matterport3d}
Chang, A., Dai, A., Funkhouser, T., Halber, M., Niessner, M., Savva, M., Song,
  S., Zeng, A., Zhang, Y.: {Matterport3D}: Learning from {RGB-D} data in indoor
  environments. International Conference on 3D Vision (3DV)  (2017)

\bibitem{chang2015shapenet}
Chang, A.X., Funkhouser, T., Guibas, L., Hanrahan, P., Huang, Q., Li, Z.,
  Savarese, S., Savva, M., Song, S., Su, H., et~al.: Shapenet: An
  information-rich 3d model repository. arXiv preprint arXiv:1512.03012  (2015)

\bibitem{cordonnier2021differentiable}
Cordonnier, J.B., Mahendran, A., Dosovitskiy, A., Weissenborn, D., Uszkoreit,
  J., Unterthiner, T.: Differentiable patch selection for image recognition.
  In: Proceedings of the IEEE/CVF Conference on Computer Vision and Pattern
  Recognition. pp. 2351--2360 (2021)

\bibitem{curless1996volumetric}
Curless, B., Levoy, M.: A volumetric method for building complex models from
  range images. In: Proceedings of the 23rd annual conference on Computer
  graphics and interactive techniques. pp. 303--312 (1996)

\bibitem{cuturi2019differentiable}
Cuturi, M., Teboul, O., Vert, J.P.: Differentiable ranking and sorting using
  optimal transport. Advances in neural information processing systems
  \textbf{32} (2019)

\bibitem{dahnert2019joint}
Dahnert, M., Dai, A., Guibas, L.J., Nie{\ss}ner, M.: Joint embedding of 3d scan
  and cad objects. In: Proceedings of the IEEE/CVF International Conference on
  Computer Vision. pp. 8749--8758 (2019)

\bibitem{dai2017scannet}
Dai, A., Chang, A.X., Savva, M., Halber, M., Funkhouser, T., Nie{\ss}ner, M.:
  Scannet: Richly-annotated 3d reconstructions of indoor scenes. In:
  Proceedings of the IEEE conference on computer vision and pattern
  recognition. pp. 5828--5839 (2017)

\bibitem{dai2017bundlefusion}
Dai, A., Nie{\ss}ner, M., Zollh{\"o}fer, M., Izadi, S., Theobalt, C.:
  Bundlefusion: Real-time globally consistent 3d reconstruction using
  on-the-fly surface reintegration. ACM Transactions on Graphics (ToG)
  \textbf{36}(4), ~1 (2017)

\bibitem{fu20213d}
Fu, H., Cai, B., Gao, L., Zhang, L.X., Wang, J., Li, C., Zeng, Q., Sun, C.,
  Jia, R., Zhao, B., et~al.: 3d-front: 3d furnished rooms with layouts and
  semantics. In: Proceedings of the IEEE/CVF International Conference on
  Computer Vision. pp. 10933--10942 (2021)

\bibitem{gelfand2005robust}
Gelfand, N., Mitra, N.J., Guibas, L.J., Pottmann, H.: Robust global
  registration. In: Symposium on geometry processing. vol.~2, p.~5. Vienna,
  Austria (2005)

\bibitem{gumeli2021roca}
G{\"u}meli, C., Dai, A., Nie{\ss}ner, M.: Roca: Robust cad model retrieval and
  alignment from a single image. arXiv preprint arXiv:2112.01988  (2021)

\bibitem{herzog2015lesss}
Herzog, R., Mewes, D., Wand, M., Guibas, L., Seidel, H.P.: Lesss: Learned
  shared semantic spaces for relating multi-modal representations of 3d shapes.
  In: Computer Graphics Forum. vol.~34, pp. 141--151. Wiley Online Library
  (2015)

\bibitem{hilaga2001topology}
Hilaga, M., Shinagawa, Y., Kohmura, T., Kunii, T.L.: Topology matching for
  fully automatic similarity estimation of 3d shapes. In: Proceedings of the
  28th annual conference on Computer graphics and interactive techniques. pp.
  203--212 (2001)

\bibitem{hou2021pri3d}
Hou, J., Xie, S., Graham, B., Dai, A., Nie{\ss}ner, M.: Pri3d: Can 3d priors
  help 2d representation learning? arXiv preprint arXiv:2104.11225  (2021)

\bibitem{hua2017shrec}
Hua, B.S., Truong, Q.T., Tran, M.K., Pham, Q.H., Kanezaki, A., Lee, T., Chiang,
  H.Y., Hsu, W., Li, B., Lu, Y., et~al.: Shrec'17: Rgb-d to cad retrieval with
  objectnn dataset. In: 10th Eurographics Workshop on 3D Object Retrieval, 3DOR
  2017. pp. 25--32. Eurographics Association (2017)

\bibitem{huang2018holistic}
Huang, S., Qi, S., Zhu, Y., Xiao, Y., Xu, Y., Zhu, S.C.: Holistic 3d scene
  parsing and reconstruction from a single rgb image. In: Proceedings of the
  European conference on computer vision (ECCV). pp. 187--203 (2018)

\bibitem{ishimtsev2020cad}
Ishimtsev, V., Bokhovkin, A., Artemov, A., Ignatyev, S., Niessner, M., Zorin,
  D., Burnaev, E.: Cad-deform: Deformable fitting of cad models to 3d scans.
  In: European Conference on Computer Vision. pp. 599--628. Springer (2020)

\bibitem{izadi2011kinectfusion}
Izadi, S., Kim, D., Hilliges, O., Molyneaux, D., Newcombe, R., Kohli, P.,
  Shotton, J., Hodges, S., Freeman, D., Davison, A., et~al.: Kinectfusion:
  real-time 3d reconstruction and interaction using a moving depth camera. In:
  Proceedings of the 24th annual ACM symposium on User interface software and
  technology. pp. 559--568 (2011)

\bibitem{izadinia2017im2cad}
Izadinia, H., Shan, Q., Seitz, S.M.: Im2cad. In: Proceedings of the IEEE
  Conference on Computer Vision and Pattern Recognition. pp. 5134--5143 (2017)

\bibitem{kim2012acquiring}
Kim, Y.M., Mitra, N.J., Yan, D.M., Guibas, L.: Acquiring 3d indoor environments
  with variability and repetition. ACM Transactions on Graphics (TOG)
  \textbf{31}(6),  1--11 (2012)

\bibitem{kuo2020mask2cad}
Kuo, W., Angelova, A., Lin, T.Y., Dai, A.: Mask2cad: 3d shape prediction by
  learning to segment and retrieve. In: European Conference on Computer Vision.
  pp. 260--277. Springer (2020)

\bibitem{kuo2021patch2cad}
Kuo, W., Angelova, A., Lin, T.Y., Dai, A.: Patch2cad: Patchwise embedding
  learning for in-the-wild shape retrieval from a single image. In: Proceedings
  of the IEEE/CVF International Conference on Computer Vision. pp. 12589--12599
  (2021)

\bibitem{li2015database}
Li, Y., Dai, A., Guibas, L., Nie{\ss}ner, M.: Database-assisted object
  retrieval for real-time 3d reconstruction. In: Computer Graphics Forum.
  vol.~34. Wiley Online Library (2015)

\bibitem{li2015joint}
Li, Y., Su, H., Qi, C.R., Fish, N., Cohen-Or, D., Guibas, L.J.: Joint
  embeddings of shapes and images via cnn image purification. ACM transactions
  on graphics (TOG)  \textbf{34}(6),  1--12 (2015)

\bibitem{maninis2022vid2cad}
Maninis, K.K., Popov, S., Niesser, M., Ferrari, V.: Vid2cad: Cad model
  alignment using multi-view constraints from videos. IEEE Transactions on
  Pattern Analysis and Machine Intelligence  (2022)

\bibitem{nan2012search}
Nan, L., Xie, K., Sharf, A.: A search-classify approach for cluttered indoor
  scene understanding. ACM Transactions on Graphics (TOG)  \textbf{31}(6),
  1--10 (2012)

\bibitem{niessner2013real}
Nie{\ss}ner, M., Zollh{\"o}fer, M., Izadi, S., Stamminger, M.: Real-time 3d
  reconstruction at scale using voxel hashing. ACM Transactions on Graphics
  (ToG)  \textbf{32}(6),  1--11 (2013)

\bibitem{ohbuchi2005shape}
Ohbuchi, R., Minamitani, T., Takei, T.: Shape-similarity search of 3d models by
  using enhanced shape functions. International Journal of Computer
  Applications in Technology  \textbf{23}(2-4),  70--85 (2005)

\bibitem{van2018representation}
Van~den Oord, A., Li, Y., Vinyals, O.: Representation learning with contrastive
  predictive coding. arXiv e-prints pp. arXiv--1807 (2018)

\bibitem{osada2002shape}
Osada, R., Funkhouser, T., Chazelle, B., Dobkin, D.: Shape distributions. ACM
  Transactions on Graphics (TOG)  \textbf{21}(4),  807--832 (2002)

\bibitem{qi2017pointnet}
Qi, C.R., Su, H., Mo, K., Guibas, L.J.: Pointnet: Deep learning on point sets
  for 3d classification and segmentation. In: Proceedings of the IEEE
  conference on computer vision and pattern recognition. pp. 652--660 (2017)

\bibitem{qi2016volumetric}
Qi, C.R., Su, H., Nie{\ss}ner, M., Dai, A., Yan, M., Guibas, L.J.: Volumetric
  and multi-view cnns for object classification on 3d data. In: Proceedings of
  the IEEE conference on computer vision and pattern recognition. pp.
  5648--5656 (2016)

\bibitem{qi2017pointnet++}
Qi, C.R., Yi, L., Su, H., Guibas, L.J.: Pointnet++: Deep hierarchical feature
  learning on point sets in a metric space. Advances in neural information
  processing systems  \textbf{30} (2017)

\bibitem{radev2002evaluating}
Radev, D.R., Qi, H., Wu, H., Fan, W.: Evaluating web-based question answering
  systems. In: LREC. Citeseer (2002)

\bibitem{ravi2020pytorch3d}
Ravi, N., Reizenstein, J., Novotny, D., Gordon, T., Lo, W.Y., Johnson, J.,
  Gkioxari, G.: Accelerating 3d deep learning with pytorch3d. arXiv:2007.08501
  (2020)

\bibitem{rusu2009fast}
Rusu, R.B., Blodow, N., Beetz, M.: Fast point feature histograms (fpfh) for 3d
  registration. In: 2009 IEEE international conference on robotics and
  automation. pp. 3212--3217. IEEE (2009)

\bibitem{simonyan2014very}
Simonyan, K., Zisserman, A.: Very deep convolutional networks for large-scale
  image recognition. arXiv preprint arXiv:1409.1556  (2014)

\bibitem{sundar2003skeleton}
Sundar, H., Silver, D., Gagvani, N., Dickinson, S.: Skeleton based shape
  matching and retrieval. In: 2003 Shape Modeling International. pp. 130--139.
  IEEE (2003)

\bibitem{tombari2010unique}
Tombari, F., Salti, S., Di~Stefano, L.: Unique signatures of histograms for
  local surface description. In: European conference on computer vision. pp.
  356--369. Springer (2010)

\bibitem{uy2020deformation}
Uy, M.A., Huang, J., Sung, M., Birdal, T., Guibas, L.: Deformation-aware 3d
  model embedding and retrieval. In: European Conference on Computer Vision.
  pp. 397--413. Springer (2020)

\bibitem{voorhees2001overview}
Voorhees, E.M., et~al.: Overview of the trec 2001 question answering track. In:
  Trec. pp. 42--51 (2001)

\bibitem{whelan2015elasticfusion}
Whelan, T., Leutenegger, S., Salas-Moreno, R., Glocker, B., Davison, A.:
  Elasticfusion: Dense slam without a pose graph. Robotics: Science and Systems
  (2015)

\bibitem{Wu2021Towers}
Wu, X., Averbuch-Elor, H., Sun, J., Snavely, N.: Towers of babel: Combining
  images, language, and 3d geometry for learning multimodal vision. In: ICCV
  (2021)

\bibitem{xie2019reparameterizable}
Xie, S.M., Ermon, S.: Reparameterizable subset sampling via continuous
  relaxations. arXiv preprint arXiv:1901.10517  (2019)

\bibitem{xie2020differentiable}
Xie, Y., Dai, H., Chen, M., Dai, B., Zhao, T., Zha, H., Wei, W., Pfister, T.:
  Differentiable top-k with optimal transport. Advances in Neural Information
  Processing Systems  \textbf{33},  20520--20531 (2020)

\bibitem{you2021canonical}
You, Y., Ye, Z., Lou, Y., Li, C., Li, Y.L., Ma, L., Wang, W., Lu, C.: Canonical
  voting: Towards robust oriented bounding box detection in 3d scenes (2021)

\bibitem{zeng20173dmatch}
Zeng, A., Song, S., Nie{\ss}ner, M., Fisher, M., Xiao, J., Funkhouser, T.:
  3dmatch: Learning local geometric descriptors from rgb-d reconstructions. In:
  Proceedings of the IEEE conference on computer vision and pattern
  recognition. pp. 1802--1811 (2017)

\bibitem{zou2019complete}
Zou, C., Guo, R., Li, Z., Hoiem, D.: Complete 3d scene parsing from an rgbd
  image. International Journal of Computer Vision  \textbf{127}(2),  143--162
  (2019)

\end{thebibliography}

\appendix
\chapter*{Appendix} 
\vspace{-1cm}
In this appendix, we show class-level evaluations corresponding to the main paper in Sec.~\ref{sec:classeval}, additional quantitative analysis in Sec.~\ref{sec:additional_quant}, network architecture details in Sec.~\ref{sec:arch}, dataset class composition in Sec.~\ref{sec:data_classes}, additional evaluation details in Sec.~\ref{sec:eval_details}, our differentiable top-$k$ layer in Sec.~\ref{sec:difftopk}, and additional details about the perceptual proxy metric in Sec.~\ref{sec:proxy}.
\vspace{-0.1cm}

\section{Class-level Evaluation}\label{sec:classeval}
\vspace{-0.2cm}
Table 1 and Table 2 show a per-class evaluation of our method and baselines on seen classes using the top-1 retrieval accuracy and IoU metrics, respectively.
Table 3 and 4 include per-class results using the same metrics on unseen classes.
We demonstrate improved performance in terms of class average and on most individual class categories.
\vspace{-0.4cm}
\begin{table}[!h]
    \centering
    \resizebox{\columnwidth}{!}{
    \begin{tabular}{|l|l|r|r|r|r|r|r|r|r|r|r|r|r|r|}
    \hline
        Method                     & chair & table & cabinet & ashcan & bookshelf & display & bathtub & sofa & bed & file & washer & class avg\\ \hline
        SHOT                       & 0.03 & 0.02 & 0.00 & 0.03 & 0.03 & 0.00 & 0.07 & 0.04 & 0.00 & 0.00 & 0.03 & 0.02 \\
        FPFH                       & 0.12 & 0.03 & 0.00 & 0.07 & 0.00 & 0.08 & 0.20 & 0.09 & 0.00 & 0.17 & 0.00 & 0.07 \\
        3D CNN                     & 0.44 & 0.26 & 0.16 & 0.23 & \textbf{0.51} & 0.37 & 0.53 & 0.17 & 0.06 & \textbf{0.5} & 0.51 & 0.34 \\
        PointNet                   & 0.39 & 0.33 & 0.25 & \textbf{0.59} & 0.49 & 0.30 & 0.20 & 0.35 & 0.48 & 0.33 & 0.17 & 0.35 \\
        JointEmbedding             & 0.52 & 0.21 & 0.21 & 0.47 & 0.29 & \textbf{0.45} & 0.40 & 0.26 & \textbf{0.52} & 0.33 & 0.29 & 0.39 \\
        \hline
        Ours, Top-K training (k=5) & \textbf{0.60} & \textbf{0.51} & \textbf{0.25} & 0.49 & 0.29 & 0.37 & \textbf{0.67} & \textbf{0.37} & 0.45 & 0.33 & \textbf{0.51} & \textbf{0.44} \\
    \hline
    \end{tabular}
    }
    \caption{Per-class top-1 retrieval accuracy on the Scan-CAD Object Similarity Dataset~\cite{dahnert2019joint} (seen classes).}
    \label{tab:comparision-seen-class}
\end{table}
\vspace{-1.5cm}
\begin{table}[!h]
    \centering
    \resizebox{\columnwidth}{!}{
    \begin{tabular}{|l|l|r|r|r|r|r|r|r|r|r|r|r|r|r|}
    \hline
        Method                     & chair & table & cabinet & ashcan & bookshelf & display & bathtub & sofa & bed & file & washer & class avg\\ \hline
        SHOT                       & 0.10 & 0.12 & 0.17 & 0.11 & 0.14 & 0.13 & 0.12 & 0.14 & 0.1 & 0.16 & 0.24 & 0.15 \\
        FPFH                       & 0.19 & 0.08 & 0.17 & 0.13 & 0.10 & 0.14 & 0.28 & 0.18 & 0.13 & 0.29 & 0.15 & 0.18 \\
        3D CNN                     & 0.46 & 0.35 & 0.56 & 0.31 & \bf{0.53} & 0.49 & \bf{0.65} & 0.39 & 0.25 & \bf{0.69} & 0.74 & 0.54 \\
        PointNet                   & 0.46 & 0.39 & 0.59 & \bf{0.48} & 0.49 & 0.50 & 0.36 & \bf{0.61} & 0.55 & 0.63 & 0.69 & 0.57 \\
        JointEmbedding             & 0.52 & 0.42 & 0.56 & 0.45 & 0.35 & 0.51 & 0.51 & 0.6 & 0.48 & 0.51 & 0.74 & 0.58\\
        
        \hline
        Ours, Top-K training (k=5) & \bf{0.54} & \bf{0.46} & \bf{0.66} & 0.47 & 0.43 & \bf{0.52} & \bf{0.65} & 0.58 & \bf{0.61} & 0.59 & \bf{0.80} & \bf{0.63} \\
    \hline
    \end{tabular}
    }
    \caption{Per-class IoU scores on the Scan-CAD Object Similarity Dataset~\cite{dahnert2019joint} (seen classes).}
\end{table}
\vspace{-1.7cm}
\begin{table}[!h]
    \centering
    \begin{tabular}{|l|r|r|r|r|r|r|r|}
    \hline
        Method & bed & file & bag & lamp & printer & class avg\\ \hline
        SHOT                       & 0.00 & 0.00 & 0.00 & 0.00 & 0.00 & 0.00\\
        FPFH                       & 0.00 & 0.00 & 0.00 & 0.00 & 0.00 & 0.00\\
        3D CNN                     & 0.03 & 0.04 & 0.01 & 0.03 & 0.04 & 0.03 \\
        PointNet                   & 0.04 & 0.00 & 0.02 & 0.01 & 0.00 & 0.01 \\
        JointEmbedding             & 0.01 & 0.01 & 0.01 & 0.00 & 0.02 & 0.01 \\
        \hline
        Ours, Top-K training (k=5) & \bf{0.11} & \bf{0.05} & \bf{0.10} & \bf{0.26} & \bf{0.12} & \bf{0.12} \\
    \hline
    \end{tabular}
    \caption{Per-class top-1 retrieval accuracy on the Scan2CAD Dataset~\cite{avetisyan2019scan2cad} (unseen classes).}
\end{table}
\vspace{-1.5cm}
\begin{table}[!h]
    \centering
    \begin{tabular}{|l|r|r|r|r|r|r|r|}
    \hline
        Method & bed & file & bag & lamp & printer & class avg\\ \hline
        SHOT                       & 0.09 & 0.15 & 0.19 & 0.07 & 0.14 & 0.13 \\
        FPFH                       & 0.09 & 0.16 & 0.15 & 0.09 & 0.16 & 0.13 \\
        3D CNN                     & 0.17 & 0.47 & 0.36 & 0.25 & 0.40 & 0.33 \\
        PointNet                   & 0.24 & 0.27 & 0.27 & 0.16 & 0.23 & 0.24 \\
        JointEmbedding             & 0.23 & 0.46 & 0.35 & 0.13 & 0.38 & - 0.31 \\
        \hline
        Ours, Top-K training (k=5) & \textbf{0.40} & \textbf{0.52} & \textbf{0.51} & \textbf{0.43} & \textbf{0.49} & \textbf{0.47} \\
    \hline
    \end{tabular}
    \caption{Per-class IoU scores on the Scan2CAD Dataset~\cite{avetisyan2019scan2cad} (unseen classes).}
\vspace{-1.2cm}
\end{table}
\newpage
\section{Additional Quantitative Analysis}
\label{sec:additional_quant}
We show results of our ablation study on unseen classes (for seen classes, see Table 3 of the main paper).
For completeness, we evaluate our method on the val set of Scan2CAD which contains seen classes to further validate the performance of our method.
Finally, Table 7 shows the performance of our proxy similarity metrics if they were applied directly to the retrieval problem. 
As detailed in Sec. 4 of the main paper, this is impractical in most settings since the memory and compute requirements for the proxy metrics are prohibitively large. 
\vspace{-0.5cm}

\begin{table}[!h]
    \centering
    \begin{tabular}{|l|l|l|l|l|l|l|l|}
    \hline
    Metric & Proxy Metric & Top1 & Top5 & Cat & IoU & IoU (Top-5) & MRR \\ \hline
    \hline
    Triplet & perceptual & 0.05 & 0.16 & 0.49 & 0.40 & 0.37 & 0.12 \\
    Embedding MSE & perceptual & 0.06 & 0.18 & 0.46 & 0.35 & 0.34 & 0.13 \\
    Top-K training (k=1) & perceptual & 0.08 & 0.21 & 0.50 & 0.42 & 0.39 & 0.15 \\ 
    Top-K training (k=5) & geometric & 0.06 & 0.19 & 0.46 & 0.40 & 0.39 & 0.13 \\ 
    Top-K training (k=5) & perceptual & 0.08 & 0.24 & 0.53 & 0.41 & 0.39 & 0.17 \\ 
    Ours, Top-K training (k=5) & combined & \textbf{0.11} & \textbf{0.28} & \textbf{0.57} & \textbf{0.46} & \textbf{0.43} & \textbf{0.19} \\ \hline
    \end{tabular}
    \caption{Ablation study for the Scan2CAD dataset (unseen classes).}
\end{table}
\vspace{-1.5cm}

\begin{table}[!h]
    \centering
    \begin{tabular}{|l|r|r|r|r|r|r|r|}
        \hline
        Method                         & Top1 & Top5 & Cat & IoU$_1$ & IoU$_5$ & MRR \\ \hline \hline
        SHOT \cite{tombari2010unique}  & 0.00 & 0.00 & 0.21 & 0.11 & 0.11 & 0.00\\ 
        FPFH \cite{rusu2009fast}       & 0.00 & 0.00 & 0.25 & 0.11 & 0.11 & 0.01\\
        3DCNN (class.)* \cite{qi2016volumetric} & 0.02 & 0.06 & 0.75 & 0.24 & 0.25 & 0.05 \\  
        PointNet* \cite{qi2017pointnet}& 0.01 & 0.04 & \textbf{0.81} & 0.22 & 0.22 & 0.04\\ 
        JointEmbedding*       & 0.02 & 0.06 & 0.73 & 0.28 & 0.26 & 0.05\\
        \hline
        Ours &  \textbf{0.09} & \textbf{0.23} & 0.79 & \textbf{0.37} & \textbf{0.34} & \textbf{0.16}\\ \hline
    \end{tabular}
    \caption{CAD retrieval to scan objects on seen classes from the validation set of Scan2CAD~\cite{avetisyan2019scan2cad}. Our differentiable retrieval achieves comparable and even improved performance without using any scan-CAD association annotations. Methods with a * require scan-CAD association supervision.}
\end{table}
\vspace{-1.5cm}

\begin{table}[!h]
    \centering
    \begin{tabular}{|l|r|r|r|r|r|r|r|}
    \hline
        Metric & Top1 & Top5 & Cat & IoU & IoU (Top-5) & MRR\\ \hline
        \hline
        %geometric (no rescale) & 0.07 & 0.21 & 0.51 & 0.47 & 0.43 & 0.14 \\ 
        geometric & 0.10 & 0.24 & 0.53 & 0.48 & 0.42 & 0.17 \\ 
        VGG-Perceptual & 0.14 & \textbf{0.37} & \textbf{0.62} & 0.49 & 0.44 & 0.25 \\ 
        combined & \textbf{0.16} & 0.35 & \textbf{0.62} & \textbf{0.52} & \textbf{0.45} & \textbf{0.26} \\ \hline
    \end{tabular}
    \caption{Comparison of the different proxy metrics on the Scan2CAD dataset (unseen classes).}
    \vspace{-1.5cm}
\end{table}

\newpage
\section{Architecture Details}\label{sec:arch}
\begin{figure}[!h]
  \centering
  \includegraphics[width=0.8\linewidth]{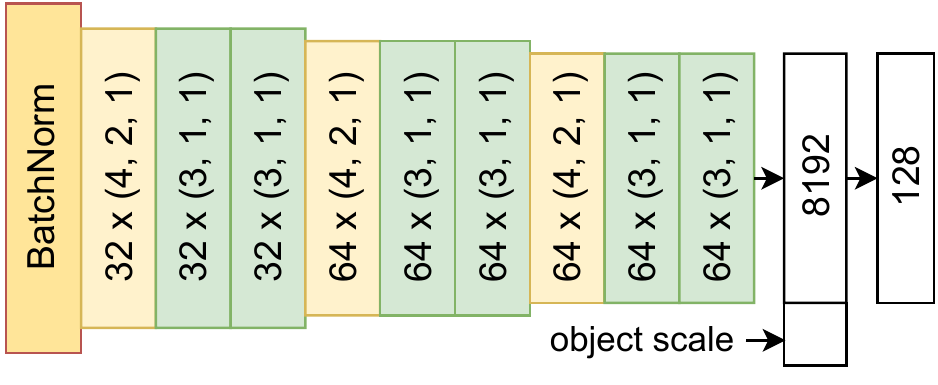}
  \caption{
  An overview of our siamese 3D-CNN based encoder.
  A CNN-block is shown as $c \times (k, s, p)$ where $c$ is the number of output channels, $k$ is the kernel size, $s$ is stride and $p$ is padding.
  Each downscale-block (yellow) consists of a convolutional layer followed by Batchnorm and a ReLU activation.
  Each residual-block (green) consists of a ResNet-style residual block.
  We concatenate the object scale to the flattened CNN feature vector before the final fully connected layer which outputs a 128-dimensional embedding vector for each object.
  }
  \label{fig:archdetail}

\end{figure}

Figure~\ref{fig:archdetail} details the architecture of our 3D CNN encoder of scan and CAD objects.

\section{Dataset Class Composition}
\label{sec:data_classes}
The dataset splits contain more than 5 instances of the following classes.
Note that the classes from Scan2CAD's test split are disjoint from those used during training.
In contrast, the Scan-CAD Object Similarity Dataset test classes are all found in the corresponding training set as well.

\begin{itemize}
    \item \textit{Scan-CAD Object Similarity Dataset (seen classes):}
        \begin{itemize}
            \item Train: \{chair, table, cabinet, ashcan, bookshelf, display, bathtub, sofa, bed, file, lamp, printer, microwave, stove, washer, basket, pot\}
            \item Val: \{chair, table, cabinet, ashcan, bookshelf, display, bathtub, sofa, bed, file\}
            \item Test: \{chair, table, cabinet, ashcan, bookshelf, display, bathtub, sofa, bed, file, washer\}
            \end{itemize}
    \item \textit{Scan2CAD (unseen classes):}
        \begin{itemize}
            \item Train: \{chair, table, cabinet, ashcan, bookshelf, display, bathtub, sofa\}
            \item Val: \{chair, table, cabinet, ashcan, bookshelf, display, bathtub, sofa\}
            \item Test: \{bed, file, bag, lamp, printer\}
        \end{itemize}
\end{itemize}

\section{Additional Evaluation Details}
\label{sec:eval_details}
\subsection{Baseline Evaluation Setup}
\paragraph{Embedding construction ablation.}
We compare a heavily optimized contrastive baseline that employs a triplet loss and positive/negative mining with a model that aims to predict soft similarities via a loss that operates on the embedding.
To select the optimal example mining strategy and triplet matches we explore multiple options via a hyperparameter search.
For example mining, we search over all valid combinations of ${\text{none}, \text{easy}, \text{semihard}, \text{hard}}$.
We try to select positives/negatives either via instance association (the matched CAD model is a positive match, all others are negatives), or via class association (the matched CAD is positive match, all models that are from a different class than the match are negatives).  
The best performing configuration for our setting selects positives/negatives via the instance association procedure and uses hard positive and negative mining.
We note that the hyperparameter configuration space is large and necessitates a time-consuming hyperparameter search to find configurations with decent performance.
To find the matched CAD model in the weakly-supervised setting, we simply select the most similar model to the scan as judged by the proxy metric.

\newpage
\section{Differentiable Top-$k$ Layer}
\label{sec:difftopk}
As described in Section~3.3 of the main paper, we use an implementation of the differentiable top-$k$ operator that was first proposed by Cordonnier et al. \cite{cordonnier2021differentiable}.
The paper contained an incorrect reference implementation, which made training impossible for our setting.
In particular, the gradients computed during the backwards pass did not consider the off-diagonal elements of the Jacobian.
For completeness, we provide a corrected version of the algorithm here. 
We emphasize that this is simply an implementation detail, not a contribution to the core ideas behind the algorithm. 

\definecolor{codegreen}{rgb}{0,0.6,0}
\definecolor{codegray}{rgb}{0.5,0.5,0.5}
\definecolor{codepurple}{rgb}{0.58,0,0.82}
\definecolor{backcolour}{rgb}{0.95,0.95,0.92}

\lstdefinestyle{mystyle}{
  backgroundcolor=\color{backcolour}, commentstyle=\color{codegreen},
  keywordstyle=\color{magenta},
  numberstyle=\tiny\color{codegray},
  stringstyle=\color{codepurple},
  basicstyle=\ttfamily\scriptsize,
  breakatwhitespace=false,
  breaklines=true,                 
  captionpos=b,                    
  keepspaces=true,                 
  numbers=left,                    
  numbersep=5pt,                  
  showspaces=false,                
  showstringspaces=false,
  showtabs=false,                  
  tabsize=2
}

\lstset{style=mystyle}
\begin{lstlisting}[language=Python, caption=PyTorch code for the Differentiable top-$k$ layer]
import torch
import torch.nn as nn

class PerturbedTopK(nn.Module):
    def __init__(self, k: int, num_samples: int = 1000, sigma: float = 0.05):
        super(PerturbedTopK, self).__init__()
        self.num_samples = num_samples
        self.sigma = sigma
        self.k = k
    def __call__(self, x):
        return PerturbedTopKFunction.apply(x, self.k, self.num_samples, self.sigma)
        
class PerturbedTopKFunction(torch.autograd.Function):
    @staticmethod
    def forward(ctx, x, k: int, num_samples: int = 1000, sigma: float = 0.05):
        b, d = x.shape
        # for Gaussian: noise and gradient are the same.
        noise = torch.normal(mean=0.0, std=1.0, size=(b, num_samples, d)).to(x.device)
        perturbed_x = x[:, None, :] + noise * sigma  # b, nS, d
        topk_results = torch.topk(perturbed_x, k=k, dim=-1, sorted=False)
        indices = topk_results.indices  # b, nS, k
        indices = torch.sort(indices, dim=-1).values  # b, nS, k
        # b, nS, k, d
        perturbed_output = torch.nn.functional.one_hot(indices, num_classes=d).float()
        indicators = perturbed_output.mean(dim=1)  # b, k, d
        # constants for backward
        ctx.k = k
        ctx.num_samples = num_samples
        ctx.sigma = sigma
        # tensors for backward
        ctx.perturbed_output = perturbed_output
        ctx.noise = noise
        return indicators
        
    @staticmethod
    def backward(ctx, grad_output):
        if grad_output is None:
            return tuple([None] * 5)
        grad_expected = torch.einsum("bnkd,bne->bkde", ctx.perturbed_output, ctx.noise)
        grad_expected /= (ctx.num_samples * ctx.sigma)
        grad_input = torch.einsum("bkde,bke->bd", grad_expected, grad_output)
        return (grad_input,) + tuple([None] * 5)
\end{lstlisting}

\newpage
\section{Perceptual Proxy Metric}
\label{sec:proxy}
\subsection{Image compositing}
\label{sec:image-compositing}
The input to the perceptual proxy metric is a composite image that consists of a normal map and a depth map for each view.
We normalize the normal map such that every RGB-pixel lies in $\left[\frac{2}{6}, \frac{4}{6}\right]\times\left[\frac{2}{6}, \frac{4}{6}\right]\times\left[\frac{2}{6}, \frac{4}{6}\right]$.
Finally, both images are blended together as follows $(1-\textbf{Depth}) * \textbf{Normals} = \textbf{Out}$

\begin{table}
\centering
    \begin{tabular}{|l | r | r | r | r | r |}
    \hline
        Azimuth & 180 & 180 & 90 & 225 & 135 \\ \hline
        Elevation & 45 & -25 & 45 & 0 & -45 \\ \hline
    \end{tabular}
    \caption{Viewpoint coordinates used to compute the perceptual proxy metric. Using a greedy search, we selected the five best performing values from the search space $\text{Azimuth} \in [0,  45,  90, 135, 180, 225, 270, 315]$ \\ $ \text{Elevation} \in [-90, -67.5, -45, -22.5,   0,  22.5,  45,  67.5,  90]$.}
    \label{tab:angles}
\end{table}
\vspace{-1.3cm}
\subsection{Sample Retrievals}
\begin{figure}[!h]
\newcommand\pimgsize{0.18\textwidth}
  \centering
  \begin{tabular}{ | c | c | c | c | c |}
    \cline{2-5}
    \multicolumn{1}{c|}{}& Scan Query & Ground Truth & Most Similar & Least Similar \\ \hline
    \parbox[t]{2mm}{\multirow{2}{*}{\rotatebox[origin=c]{90}{seen classes \;\;\;\;}}}&
    \begin{minipage}{\pimgsize}
      \includegraphics[width=\linewidth]{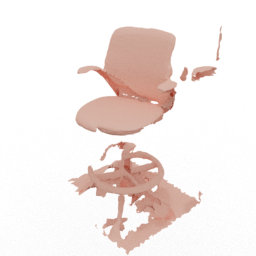}
    \end{minipage}
    &
    \begin{minipage}{\pimgsize}
      \includegraphics[width=\linewidth]{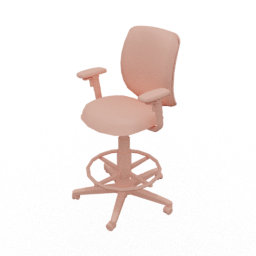}
    \end{minipage}
    & 
    \begin{minipage}{\pimgsize}
      \includegraphics[width=\linewidth]{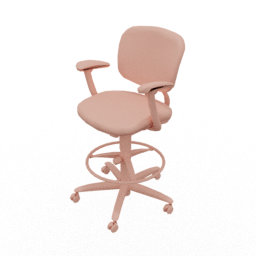}
    \end{minipage}
    & 
    \begin{minipage}{\pimgsize}
      \includegraphics[width=\linewidth]{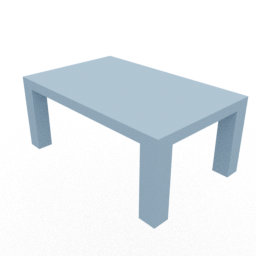}
    \end{minipage}
    \\ \cline{2-5} 
    &
    \begin{minipage}{\pimgsize}
      \includegraphics[width=\linewidth]{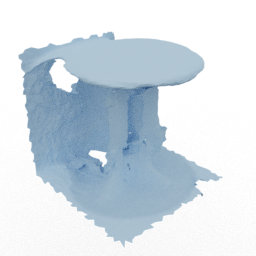}
    \end{minipage}
    &
    \begin{minipage}{\pimgsize}
      \includegraphics[width=\linewidth]{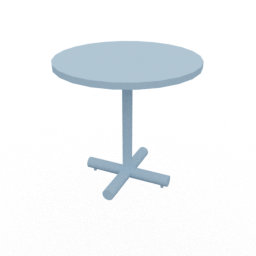}
    \end{minipage}
    & 
    \begin{minipage}{\pimgsize}
      \includegraphics[width=\linewidth]{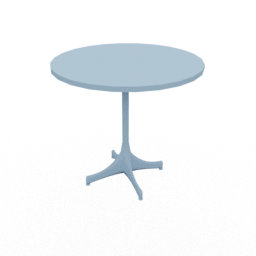}
    \end{minipage}
    & 
    \begin{minipage}{\pimgsize}
      \includegraphics[width=\linewidth]{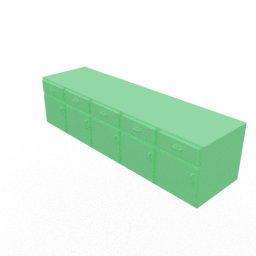}
    \end{minipage} 
    \\ \hline   \hline
    \parbox[t]{2mm}{\multirow{2}{*}{\rotatebox[origin=c]{90}{unseen classes \;\;\;\;}}}&
    \begin{minipage}{\pimgsize}
      \includegraphics[width=\linewidth]{media/Proxy Diagram/scene0557_02__2.jpg}
    \end{minipage}
    &
    \begin{minipage}{\pimgsize}
      \includegraphics[width=\linewidth]{media/Proxy Diagram/e91c2df09de0d4b1ed4d676215f46734.jpg}
    \end{minipage}
    & 
    \begin{minipage}{\pimgsize}
      \includegraphics[width=\linewidth]{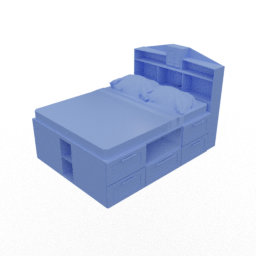}
    \end{minipage}
    & 
    \begin{minipage}{\pimgsize}
      \includegraphics[width=\linewidth]{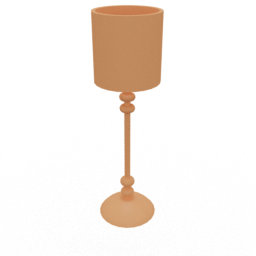}
    \end{minipage}
    \\ \cline{2-5}
    &
    \begin{minipage}{\pimgsize}
      \includegraphics[width=\linewidth]{media/Proxy Diagram/scene0024_00__19.jpg}
    \end{minipage}
    &
    \begin{minipage}{\pimgsize}
      \includegraphics[width=\linewidth]{media/Proxy Diagram/d8d129b1a07b23b7a738de48265832af.jpg}
    \end{minipage}
    & 
    \begin{minipage}{\pimgsize}
      \includegraphics[width=\linewidth]{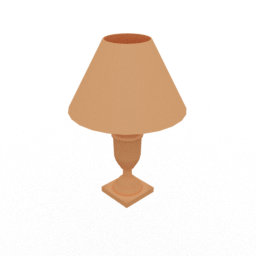}
    \end{minipage}
    & 
    \begin{minipage}{\pimgsize}
      \includegraphics[width=\linewidth]{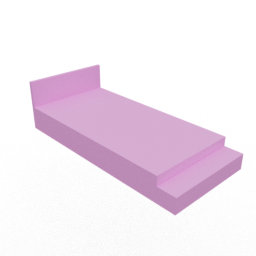}
    \end{minipage}
    \\ \hline
  \end{tabular}
  \vspace{-0.2cm}
  \caption{Qualitative illustration of the performance of our combined proxy similarity metric. 
  Top: Query scans and retrievals from the test-set of the scan-CAD-object similarity dataset. Bottom: Retrievals from the Scan2CAD validation set.
  }\label{fig:proxy-metric}
  \vspace{-0.7cm}

\end{figure}

Fig.~\ref{fig:proxy-metric} shows example CAD retrievals directly using the combined proxy metric for retrieval. 
As mentioned in Sec.~\ref{sec:additional_quant}, this is impractical for most applications due to the compute and memory requirements. 
\vspace{-0.4cm}

\end{document}